\documentclass{article} 
\usepackage{arxiv/iclr2025_conference,times}

\usepackage{amsmath,amsfonts,bm}

\def\eqref#1{equation~\ref{#1}}

\def\1{\bm{1}}

\DeclareMathAlphabet{\mathsfit}{\encodingdefault}{\sfdefault}{m}{sl}
\SetMathAlphabet{\mathsfit}{bold}{\encodingdefault}{\sfdefault}{bx}{n}

\usepackage{hyperref}
\usepackage{url}

\usepackage[pdftex]{graphicx} 
\usepackage{subcaption}       
\usepackage{siunitx}          
\usepackage{svg} 
\usepackage{longtable}
\usepackage{booktabs}       
\usepackage{amsfonts}       
\usepackage{nicefrac}       
\usepackage{amsmath}
\usepackage{xcolor}
\definecolor{thecolor}{RGB}{46, 95, 127}
\usepackage[utf8]{inputenc}
\usepackage[pdftex]{graphicx} 
\usepackage{subcaption}       
\usepackage{siunitx}          
\usepackage[T1]{fontenc}
\usepackage{microtype}      

\newcommand{\taskname}{SORT\ }
\newcommand{\datasetname}{Book-SORT\ }

\author{Mathis Pink$^1$, Vy Ai Vo$^2$, Qinyuan Wu$^1$, Jianing Mu$^3$, Javier Turek$^2$, Uri Hasson$^{4,5}$, \\
  \textbf{Kenneth A. Norman$^{4,5}$, Sebastian Michelmann$^6$, Alexander Huth$^3$, Mariya Toneva$^1$} \\
  $^1$Max Planck Institute for Software Systems, Saarbrücken, Germany \\
  $^2$Intel Labs, Hillsboro, Oregon \\
  $^3$Department of Computer Science, University of Texas at Austin, Texas \\
  $^4$Department of Psychology, Princeton University, Princeton, New Jersey \\
  $^5$Princeton Neuroscience Institute, Princeton University, Princeton, New Jersey \\
  $^6$Department of Psychology, New York University, New York City, New York \\
  \texttt{\{mpink, qwu, mtoneva\}@mpi-sws.org}\\
  \texttt{\{vy.vo, javier.turek\}@intel.com}\\
  \texttt{\{hasson, knorman\}@princeton.edu}\\
  \texttt{jmu@utexas.edu, huth@cs.utexas.edu}\\
  \texttt{s.michelmann@nyu.edu}
}

\newcommand\blfootnote[1]{%
  \begingroup
  \renewcommand\thefootnote{}\footnote{#1}%
  \addtocounter{footnote}{-1}%
  \endgroup
}

\iclrfinalcopy
\begin{document}
\title{Assessing Episodic Memory in LLMs \\ with Sequence order recall tasks}
\maketitle

\begin{abstract}

Current LLM benchmarks focus on evaluating models' memory of facts and semantic relations, primarily assessing semantic aspects of long-term memory. However, in humans, long-term memory also includes episodic memory, which links memories to their contexts, such as the time and place they occurred. The ability to contextualize memories is crucial for many cognitive tasks and everyday functions. This form of memory has not been evaluated in LLMs with existing benchmarks. To address the gap in evaluating memory in LLMs, we introduce Sequence Order Recall Tasks (SORT), which we adapt from tasks used to study episodic memory in cognitive psychology. SORT requires LLMs to recall the correct order of text segments, and provides a general framework that is both easily extendable and does not require any additional annotations. We present an initial evaluation dataset, Book-SORT, comprising $36$k pairs of segments extracted from $9$ books recently added to the public domain. Based on a human experiment with $155$ participants, we show that humans can recall sequence order based on long-term memory of a book. We find that models can perform the task with high accuracy when relevant text is given in-context during the SORT evaluation. However, when presented with the book text only during training, LLMs' performance on SORT falls short. By making it possible to evaluate more aspects of memory, we believe that SORT will aid in the emerging development of memory-augmented models.
\blfootnote{Code: \href{https://github.com/bridge-ai-neuro/SORT}{https://github.com/bridge-ai-neuro/SORT}}
\blfootnote{Dataset: \href{https://huggingface.co/datasets/memari/booksort}{https://huggingface.co/datasets/memari/booksort}}
\end{abstract}

\section{Introduction}

Large language models (LLMs) have impressive performance on many benchmarks that test factual or semantic knowledge learned during training or in-context \citep{hendrycks2020measuring,ryo2023wice, logan2019barack,petroni2019language,yu2023kola,sun2023head}.
While these advances are noteworthy, the type of long-term knowledge that these datasets test is only one of several types that naturally intelligent systems store, retrieve, and update continuously over time \citep{norris2017short,izquierdo1999separate,mcclelland1995there}. Current evaluation tasks do not assess episodic memory, which is a form of long-term knowledge thought to be important for cognitive function in humans and animals. In contrast to semantic memory, episodic memory links memories to their contexts, such as the time and place they occurred. This ability to organize memory based on spatial and temporal details enables us to reconstruct events that occurred in the possibly distant past, predict the future, and relate information across multiple events that are separated by time windows spanning a lifetime, capabilities crucial for many cognitive tasks and everyday functions.

The ability to link temporal context to stored information may be key to improving LLM performance on several tasks. More human-like episodic memory may improve models' continual learning and adaptation to shifting data distributions, performance on tasks requiring long contexts (e.g., long chat exchanges with a user), and source attribution via knowledge of where and when a memory was acquired, which could help to reduce or identify hallucinations.

To address the gap in evaluating memory in LLMs, we propose the Sequence Order Recall Task (SORT), which we adapt from tasks in cognitive psychology that are used to assess long-term episodic memory in humans and animals \citep{eichenbaum2013memory,davachi2015hippocampus}. Specifically, SORT requires a model to recall the correct order of sequential data, such as segments of text. 

We provide a specific instantiation of SORT that requires models to recall the correct order of two segments sampled from text, along with a corresponding evaluation dataset--Book-SORT. Book-SORT contains over $36$k pairs of text segments from $9$ books, with variations in segment length ($20$ and $50$ words) and distance between segments (up to $16$k words). We chose books that were very recently released from U.S. copyright to minimize the possibility that LLMs were pre-trained on these texts. 
This allowed us to test three common methods of giving a language model access to a specific text: (1) during inference in-context, (2) during inference via retrieval augmented generation (RAG), and (3) during training via fine-tuning with a language modeling objective. Furthermore, we provide a human evaluation from $155$ participants who had finished reading a whole book and were tested with no additional access to the book, showing that humans can recall segment order with up to $70\%$ accuracy based on their long-term memory of the book. While the ceiling performance on SORT is $100\%$ (assuming that texts do not contain duplicate segments), our human data provides an important reference point to compare and contrast long-term memory across models and humans.

When given access to excerpts from the books in-context, we find that models achieve up to $95\%$ accuracy with relevant $250$-word excerpts but degrade quickly as longer excerpts are presented. When models use RAG instead, they can recall sequence order only with limited performance below $65\%$. Finally, models fine-tuned with a language modeling objective on the book texts do not significantly improve their SORT performance, showing that parametric memory in current transformer models supports semantic but not episodic long-term memory.

Our main contributions can be summarized as follows:
\begin{itemize}
	\item proposal of the self-supervised task SORT, which requires LLMs to recall the correct order of segments from a sequence and can be used to assess capabilities in LLMs that would be supported by episodic memory in humans
	\item a new dataset \datasetname composed of $36$k samples from $9$ public domain books and an evaluation framework that is easily extendable to new datasets
	\item first-of-its-kind human evaluation ($N=155$) 
	showing that humans are capable of recalling the order of text from an entire book based on long-term memory
	\item a comprehensive evaluation of open-source and closed language models on Book-SORT, showing that current models: i) have good in-context memory performance, when all necessary information is presented in the prompt and the prompt is short; ii) quickly lose the ability to recall sequence order as the excerpt provided in-context gets longer, even though the excerpt still easily fits within the context window; (iii) fail to recall segment order based on parametric memory formed via fine-tuning with a language modeling objective; (iv) perform worse on SORT with retrieval augmented memory than with in-context memory.
\end{itemize}

\section{Related Work}

\textbf{Evaluation of parametric semantic memory in LLMs.}
Benchmarks such as MMLU \citep{hendrycks2020measuring}, T-REx \citep{elsahar2018t}, LAMA \citep{petroni2019language}, WICE \citep{ryo2023wice}, KoLA \citep{yu2023kola}, and others \citep{sun2023head} test models' retrieval and reasoning ability on different domains, such as recalling a chemistry fact. 

Other benchmarks that partially evaluate LLM semantic memory are those that require reasoning using temporal \citep{Ning2020TORQUEAR,zhou2021temporal,feng2023generic} (e.g. lunch happens before dinner), causal \citep{srivastava2023beyond} (e.g. she is eating, therefore she is hungry), or other commonsense knowledge (e.g. food is edible) \citep{ismayilzada2023crow} acquired during pretraining. In contrast to these benchmarks, our work proposes a task that involves judgments regarding temporal context information about text segments that either (a) are available through in-context memory or (b) were otherwise previously presented to the model, e.g. via fine-tuning or Retrieval Augmented Generation, and is agnostic of the specific semantic content of these segments.

\textbf{Evaluation of in-context memory in LLMs.} Among other conditions, we evaluate in-context memory, in which the model has in-context access to all relevant text for the task. This relates to works that evaluate a model's ability to reason over its context input, such as Needle In A Haystack \citep{LLMTest_NeedleInAHaystack} and FLenQA \citep{levy2024same}. 

Previous datasets and benchmarks that evaluate performance over long context lengths, such as Long Range Arena \citep{tay2021long}, SCROLLS \citep{shaham2022scrolls}, and MULD \citep{hudson2022muld}, are also relevant.
The evaluation of in-context memory with SORT differs from these works by focusing on order information, which is key to episodic memory in humans. Additionally, we use SORT to evaluate parametric memory which contains information beyond the current context.

\textbf{Tasks related to SORT.} Previously proposed tasks that most closely relate to SORT are BART's denoising training objective \citep{lewis2020bart}, which permutes the order of sentences in a document and learns to reconstruct the correct order, and BERT's next sentence prediction objective \citep{devlin2019bert}, which learns to predict whether two sentences follow each other in a text. SORT differs from these tasks, as it is not intended as a training objective, and it can include text segments with an arbitrary distance between each other in a document, possibly exceeding the context input length of the model. 
In ChapterBreak \citep{sun-etal-2022-chapterbreak}, long segments ending at a chapter boundary taken from a book are presented to an LLM along with multiple segments of chapter beginnings from the same book. The task for the LLM is then to tell which one is the directly following chapter and which are not. This suffix-identification task aims to evaluate narrative-understanding based reasoning about books, while we propose SORT as an evaluation for episodic memory in LLMs, involving both a model and a memory-insertion method. By evaluating a SORT baseline in which the models do not have access to relevant source texts, we show that memory is needed for SORT and general narrative-reasoning ability is not enough.

\section{Sequence Order Recall Task}
\label{sec:sort_methods}

\begin{figure}[t]
    \centering
    \includegraphics[width=1.0\textwidth]{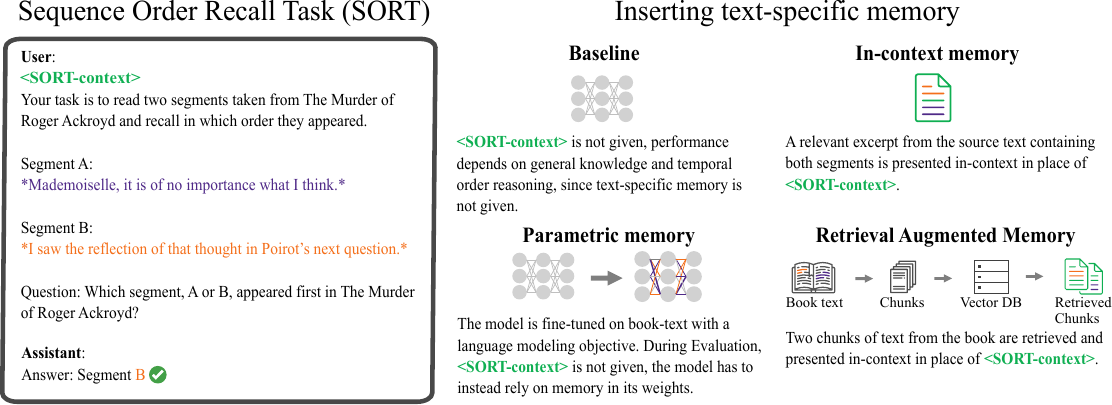}
    \caption{Overview of the Sequence Order Recall Task (SORT) to evaluate how models can access memory of temporal order. Left: Example task prompt for SORT. A prefix to the prompt can be given to assess in-context forms of memory. Right: Methods to insert memory of specific texts into a model.}
    \label{fig:sort-overview}
\end{figure}

We introduce a novel evaluation task: recalling the order of parts of a sequence, which we term the Sequence Order Recall Task (SORT). SORT is adapted from recency judgment tasks used in cognitive psychology to evaluate episodic memory in humans and animals \citep{eichenbaum2013memory,davachi2015hippocampus}. In this task, a sequence is presented to a participant. Then, after some delay, the participant is asked to judge the order in which two segments of the sequence appeared. We adapt this task to test memory in models. The general task can be applied to any sequential domain, including video and audio. Here we focus on the text domain to evaluate LLMs (Fig. \ref{fig:sort-overview}).

\textbf{Formal description of SORT.} 
The general form of the task can be described as follows. Let $\mathbf{X} \in \mathbb{R}^{T \times F}$ be sequential data, where $\bf{T}$ is the number of time-steps (e.g.\ token in a text) and $\bf{F}$ is the number of features (e.g.\ vocabulary size). We define start indices $\mathbf{t_{j}}$ and $\mathbf{t_{k}}$ for pairs of segments of length $\mathbf{L} \in \mathbb{N}^+$ in $\mathbf{X}$, such that both $\mathbf{t_j < t_{k}}$ and $\mathbf{t_j+L \leq t_{k}}$. Using these, we extract non-overlapping segments from the original sequence $\mathbf{X}$ as $\mathbf{\widetilde{X}_i = X[t_i : t_i + L-1, :]}$. 
The order of segments $\mathbf{\widetilde{X}_j}$ and $\mathbf{\widetilde{X}_{k}}$ is randomized, yielding $\mathbf{[\widetilde{X}_A \; \widetilde{X}_B]}$, which is then given as part of a model's input. The task for a model $\mathbf{\mathcal{M}_\theta}$ is to infer whether $\mathbf{t_A < t_B}$, i.e.\ in SORT, the task of a model is to predict which of two non-overlapping subsequences $\mathbf{\widetilde{X}_A}$ and $\mathbf{\widetilde{X}_B}$ has the lower starting index in $\mathbf{X}$. The task can be used to evaluate a variety of methods to include document-specific memory in models. To assess in-context memory, i.e. memory based on text presented in-context, the segments are preceded by $\mathbf{X}$ in the model's input. When assessing retrieval-augmented generation methods, instead of prepending $\mathbf{X}$, passages of $\mathbf{X}$ are retrieved and prepended. For the assessment of parametric long-term memory, $\mathbf{X}$ is not part of a model's input, instead the model's parameters $\mathbf{\theta}$ are a function of $\mathbf{X}$ via pre-training or fine-tuning: $\theta = f(\mathbf{X})$.

The general form of SORT is the following input, which can be preceded by additional context to insert a memory:
\begin{align}
I_{SORT} = [P_{context} \; P_{task} \; P_{label_A} \; \mathbf{\widetilde{X}_A}\;  P_{label_B} \; \mathbf{\widetilde{X}_B} \; P_{question} P_{answer}],
\label{eq:input_def}
\end{align}
where $\bf{P_{context}}$ can either be relevant context, such as (parts of) the source sequence $\mathbf{X}$ to assess in-context memory (stored in activation slots), or an empty string when parametric memory (stored in weights) is assessed; $\bf{P_{task}}$ instructs the model for the sequence order recall task to read two segments and describes the objective: answering which of the two labeled segments appears first in $\mathbf{X}$; $\bf{P_{label_A}}$ and $\bf{P_{label_B}}$ are the labels (e.g.\ the characters ``A'' and ``B'') for the first and second segment presented in the task $\mathbf{\widetilde{X}_A}$ and $\mathbf{\widetilde{X}_B}$; $\bf{P_{question}}$ repeats the SORT objective as a question; finally, $\bf{P_{answer}}$ provides the beginning of the answer string as ``Answer: Segment''.

\subsection{Evaluating Large Language Models on sort}
\label{subsec:eval_sort}

We greedily sample an answer token $\mathbf{a = argmax(\mathcal{M}_\theta(I))}$ from the model $\mathbf{\mathcal{M}}_\theta$, which is parameterized by $\mathbf{\theta}$, and decode the sampled answer token $\mathbf{a}$ as either "A" or "B".

The answer is evaluated as correct if it corresponds to the segment that truly appears first in $\mathbf{X}$.
For proprietary (OpenAI) models that do not allow completing assistant responses with prepended text, we omit $\bf{P_{answer}}$. In this case we resort to generating a sequence of 25 tokens, and parse the generated text for A or B responses.

\textbf{Prompt selection.} Using a single prompt formulation across all models may bias the results. To prevent this, we compiled a set of $12$ prompts that vary formulations in $\bf{P_{context}}$ and $\bf{P_{task}}$. For each model, we evaluate each prompt on a held-out dataset of 400 samples and used the best performing prompt for each model. The full prompts and further details on prompt selection are given in Appendix \ref{appendix:prompts}-\ref{appendix:prompts-results}. 

\textbf{Baseline without book-specific memory.} We want to ensure that performance on SORT is due to text-specific memory and not due to temporal order reasoning supported by more semantic forms of memory such as commonsense knowledge (e.g. lunch happens before dinner). We isolate the effects on SORT that are due to text-specific memory by contrasting performance between a baseline model that does not have access to the specific text and a model that has access to the sequences in one of various ways in which memory can be inserted.

\subsection{Inserting text-specific memory into models}
\label{subsec:memory_methods}

We evaluate three methods to insert text-specific memory into models: (1) via in-context presentation, (2) via fine-tuning with a language modeling objective, and (3) via retrieval augmented generation of short chunks of text in a book.

\textbf{In-context presentation.} When assessing in-context memory, $\bf{P_{context}}$ in Eq. \ref{eq:input_def} contains relevant excerpts from the source text along with the book title. The prompt includes the instruction to carefully read the text from the book (a list of used prompts is shown in Appendix \ref{appendix:prompt-list}). To test in-context memory, We make sure that excerpts contain both segments and vary the length of excerpts in our experiments. 

\textbf{Finetuning with a language modeling objective.} Instead of presenting text from the books in the same prompt in which the SORT task is given, we are interested in parametric memory of the texts. In this condition, $\bf{P_{context}}$ in Eq. \ref{eq:input_def} is an empty string. To insert parametric memory of the source texts into a model, we fine-tune the model with a next-token prediction objective on the books, split into chunks of 5000 words and contextualized by the books' titles. Since we need to preserve the models' ability to understand and follow the task instructions, we fine-tune on a dataset that additionally includes 3,500 random instruction-following examples that are unrelated to SORT. This helps to prevent catastrophic forgetting during continued finetuning \citep{luo2024empiricalstudycatastrophicforgetting}. We finetune on 8 A100 GPUs with an initial learning rate of 5e-6 and a batch size of 192. Full details of the fine-tuning setup are given in Appendix \ref{appendix:finetuning} and our code will be available. 

\textbf{Retrieval Augmented Generation.}
To include memory of text via retrieval augmented generation (RAG), we built a typical naive RAG pipeline that relies on two separately pretrained models for the retriever and the reader \citep{gaoRetrievalAugmentedGenerationLarge2024}. The retriever returns text passages from a database to serve as task context for the LLM (i.e. as $\bf{P_{context}}$, Eq. \ref{eq:input_def}). 

The retrieval database contained text embeddings of all passages from \datasetname (Sec. \ref{sec:booksort}). We used the LangChain recursive text splitter to chunk \datasetname text into $\sim$1024 character, non-overlapping passages (average 183 words). Each passage was then encoded into a 1024-d vector using a high-performing, open-source text retrieval model (BGE-v1.5, \cite{xiaoCPackBGE2024}). To retrieve the passages, we used the Faiss \citep{douzeFaissLibrary2024} library to conduct an exact nearest neighbor search. The search returned the $k=2$ nearest neighbors. We maintained this similarity order when inserting the retrieved passages into the prompt, i.e. the most similar passage appears first in $\mathbf{P_{context}}$. As described in Section \ref{subsec:eval_sort}, we selected a single prompt for each model based on the model's performance on the held-out validation set across 10 different possible prompts (see Appendix \ref{appendix:prompts-rag}).

\section{Book-sort Dataset and Evaluation}
\label{sec:booksort}

We created an English language dataset to evaluate episodic memory in humans and LLMs. The selected sequence data considered several factors: (1) we chose long texts (mean length = 72,700 words) that exceed the context windows of most transformer LLMs; (2) we used books to enhance memorability for human readers and facilitate our human evaluation experiment; (3) we selected books from \emph{Project Gutenberg} that recently entered the U.S.\ public domain to avoid ethical and copyright issues, and minimize pre-training contamination in LLMs. Within these constraints, we aimed to maximize content diversity, including narrative fiction novels, a physics text, and an extended essay. Further details on the $9$ books in the \datasetname dataset are available in Appendix \ref{appendix:books}.

\subsection{\datasetname Creation}

We constructed a dataset that varies across factors that can affect human or model performance on SORT. Based on prior reports on LLMs \citep{liu2024lost}, we first varied (1) $L_E$, the length of the text excerpt presented in context. Since the typical standard context length of the LLMs in our study was 4096 tokens, we set $L_E=\{250, 1000, 2500\}$ words. For models with extended context windows, we also created datasets where $L_E=\{10000, 20000\}$ words, which excluded one book that was too short. Our pilot experiments on humans suggested two other factors that would affect task performance: (2) $L_S$, the length of the segments from the text, and (3) $D_S$, the distance between the segments in the original text. To mirror the human experiments, we set $L_S=\{20, 50\}$ words. We then created 4 different distance bins $D_S=\{d_0, d_1, d_2, d_3\}$, whose values were bounded by the excerpt length $L_E$ (Appendix Table \ref{tab:segdist}).

Within each unique combination of the first two factors $L_E$ and $L_S$, we randomly sampled 110 excerpts from each of the 9 books (i.e.\ 100 samples for SORT evaluation, and 10 samples for prompt selection per book). All excerpts and segments began at a sentence boundary. Within each combination of $L_E, L_S$, we randomly sampled 4 different segment pairs, one from each distance bin $D_S$. This minimized the possibility that observing an effect of distance on \taskname performance would be due to differences in the semantic content of the text segments. Finally, for all 110 trials within each of these 3 factors, we counterbalanced the correct answer. This yielded a well-controlled and easily extendable dataset of about $36K$ text segment pairs for \taskname evaluation.

\subsection{Human long-term memory evaluation}

As a reference point (but not a performance ceiling), we further provide a human evaluation from $155$ participants who had recently finished reading one of the $9$ books in the \datasetname dataset, \emph{The Murder of Roger Ackroyd} \citep{christie1927murder}. This evaluation assessed long-term memory, as the average time between reading and testing was $7.5$ days, far surpassing short-term memory duration \citep{hasson2015hierarchical}. There is no previously reported data on long-term memory for entire books from large samples, so we designed an experiment to collect this data. Given the difficulty of recruiting participants to read lengthy books specifically for an experiment, we used a creative recruiting strategy: inviting members of the online reading community \emph{Goodreads} who had recently finished \emph{The Murder of Roger Ackroyd}.  Participants completed an online survey within $30$ days of finishing the book. The expected compensation for participation was $\$12$ and the study was approved by the IRB at Anonymized University. We provide 1570 segment pair samples from 155 participants. Further details about this one-of-a-kind study are provided in Appendix \ref{appendix:human_exp}.

\subsection{Models}

We evaluate a selection of open models covering a broad range of scores on popular benchmarks such as MMLU (see Table \ref{table:all_models_name}) ranging from 7b to 8x22b parameter transformer models. Initial experiments with non-instruction-tuned models resulted in chance performance on Book-SORT (see Appendix \ref{appendix:more-models}), which we attribute to the lack of instruction tuning\footnote{ \citep{zhang2024instruction} provides an overview of instruction tuning approaches}, and thus focus on evaluating instruction-tuned models in this work. We have selected models from different model families including Llama3 \citep{llama3modelcard}, Llama2 \citep{touvron2023llama}, Mistral \citep{jiang2023mistral}, Mixtral \citep{jiang2024mixtral}, Gemma \citep{team2024gemma} and OpenAI GPTs \citep{achiam2023gpt}. For our experiments on finetuning as a method for inserting memory into models, we focus on two models Mistral-v0.2-7b-Instruct and Llama3-8b-Instruct because they allow full-parameter fine-tuning with 8 A100 GPUs.

\section{Results}
\label{sec:results}
We present empirical findings for a baseline without text-specific memory of the books in Book-SORT, as well as three methods to include memory, using 9 open-source models and 2 closed language models.

\subsection{Baseline}

\textbf{SORT requires memory specific to books in Book-SORT.} To validate that it is not possible to achieve high performance on Book-SORT without memory of the specific books that are included in the dataset, we evaluate models before they have access to the books. We find that segment pairs with a very short and with a very long distance in the book allow for above-chance-performance (see Appendix \ref{appendix:baseline}), indicating that some of these segment pairs can be ordered based not on memory but rather on temporal-order reasoning or common-sense. However, performance is below 60\% for all models and segment lengths, confirming that SORT requires memory for the particular books being queried to yield high levels of performance. 

\begin{table}[t]
\begin{center}
\captionsetup{skip=1\baselineskip} 
\caption{Baseline: SORT performance before models are exposed to the books in Book-SORT.}
\label{tab:ltm performance}
\begin{tabular}{c c c} 
\hline
\textbf{} & \textbf{Segment length 20} & \textbf{Segment length 50} \\
\hline
Llama3-70b-inst & $0.52\pm0.007$ & $0.54\pm0.007$ \\
Llama3-8b-inst & $0.51\pm0.008$ & $0.52\pm0.007$ \\
Mixtral-8x22b-inst & $0.52\pm0.007$ & $0.55\pm0.007$\\
Mixtral-8x7b-DPO-inst & $0.52\pm0.008$ & $0.54\pm0.008$ \\
Llama2-70b-inst & $0.51\pm0.007$ & $0.51\pm0.008$ \\
Gemma-1.1-7b-inst & $0.51\pm0.008$ & $0.51\pm0.007$ \\
Mistral-v0.2-7b-inst & $0.51\pm0.007$ & $0.51\pm0.008$\\
Mistral-v0.1-7b-inst & $0.50\pm0.008$ & $0.50\pm0.008$\\ 
Llama2-7b-inst & $0.50\pm0.008$ & $0.49\pm0.008$\\
GPT-3.5-turbo & $0.52\pm0.009$ & $0.52\pm0.012$\\
GPT-4 & $0.53\pm0.008$ & $0.57\pm0.007$\\
 \hline
\end{tabular}
\end{center}
\end{table}

\begin{figure}[t]
    \centering
 \includegraphics[width=0.8\linewidth]{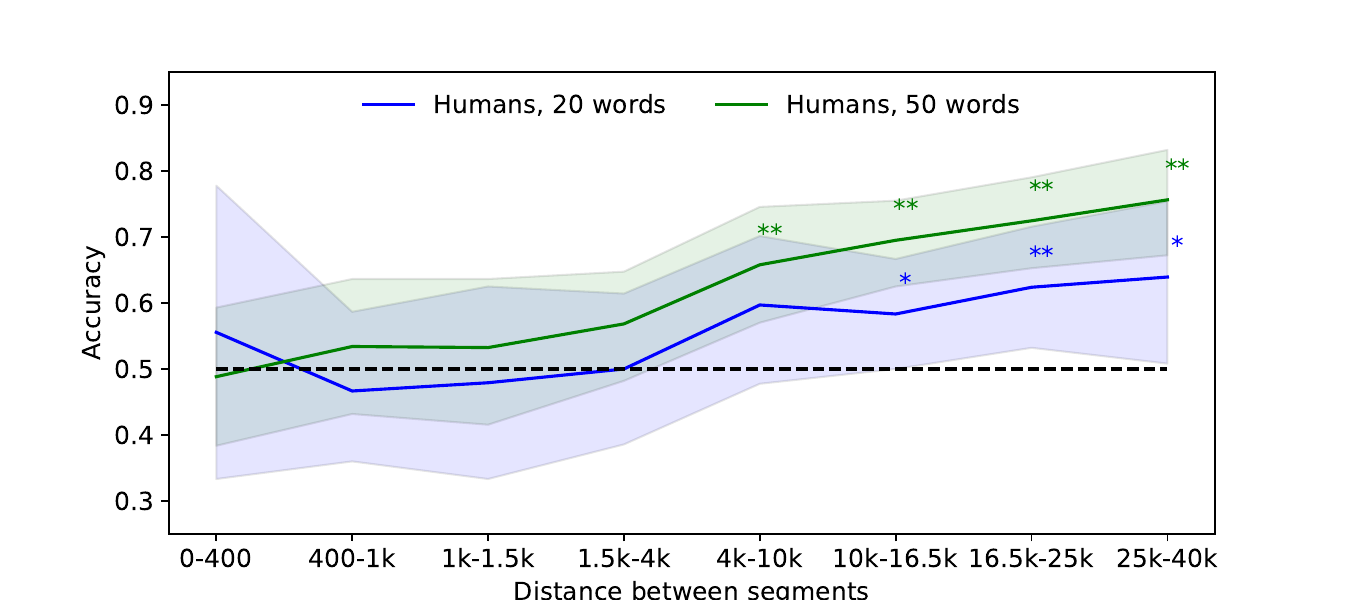}
    \caption{Human long-term memory performance on \taskname for different segment lengths and distances between segments. Shaded areas depict bootstrapped 95\% confidence intervals. Significant difference from chance is marked with asterisks ($^*$p-value$<$0.05,$^{**}$p-value$<$0.01).}
    \label{fig:humans-LTM}
\end{figure}

\subsection{Human Experiment}

\textbf{Humans can perform in SORT based on long-term memory.} The results from human long-term memory (LTM) experiments, depicted in Figure \ref{fig:humans-LTM}, demonstrate that humans can perform in SORT based on long-term memory. The average accuracy is $0.64$ for segments of 50 words and $0.56$ for segments of 20 words).
Human performance is higher for pairs of segments that have a greater distance in the book, with a peak accuracy of $0.76$ for distances greater than 25,000 words and 50-word segments. Binomial tests show that beyond a distance of 4000 words, humans perform statistically significantly better than chance. Note that we present these results as evidence that one possible information processing system--a human--can perform SORT based on long-term memory. Importantly, these results do not present the ceiling performance on the memory task that we propose. The expected ceiling performance on SORT is 100\%, assuming that the books do not contain duplicated segments of text; the odds of exact duplication decrease as segment length increases.

\subsection{In-context memory}
\textbf{Models generally perform well on SORT based on in-context memory.} Nearly all models achieve above 77\% accuracy when given in-context access to relevant excerpts from the books, reaching up to 95\% (Table \ref{tab:stm_results}). This indicates that very large models are not necessary to perform this task effectively, as demonstrated by the Llama3-8b model outperforming larger models such as Llama3-70b and Mixtral-8x7b-DPO.

\begin{figure}[t]
    \begin{subfigure}[t]{0.5\textwidth}
    \includegraphics[width=\textwidth]{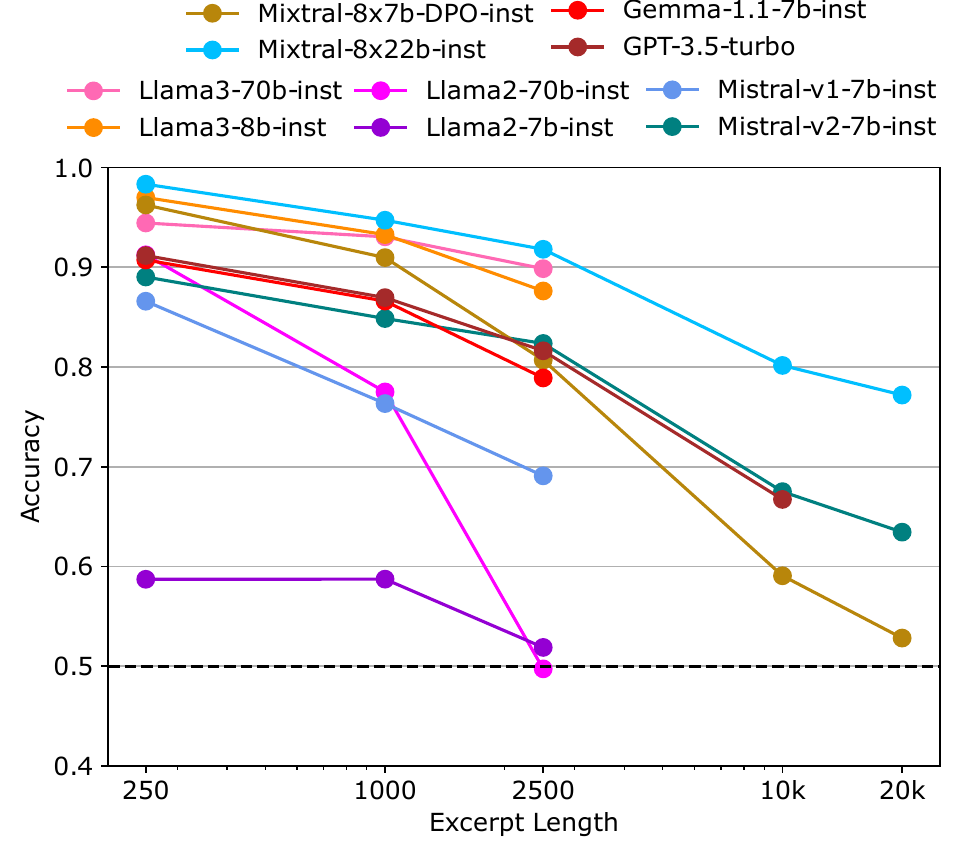}
        \caption{By excerpt lengths}
        \label{fig:excerpt_length}
    \end{subfigure}
    \begin{subfigure}[t]{0.495\textwidth}
    \includegraphics[width=\textwidth]{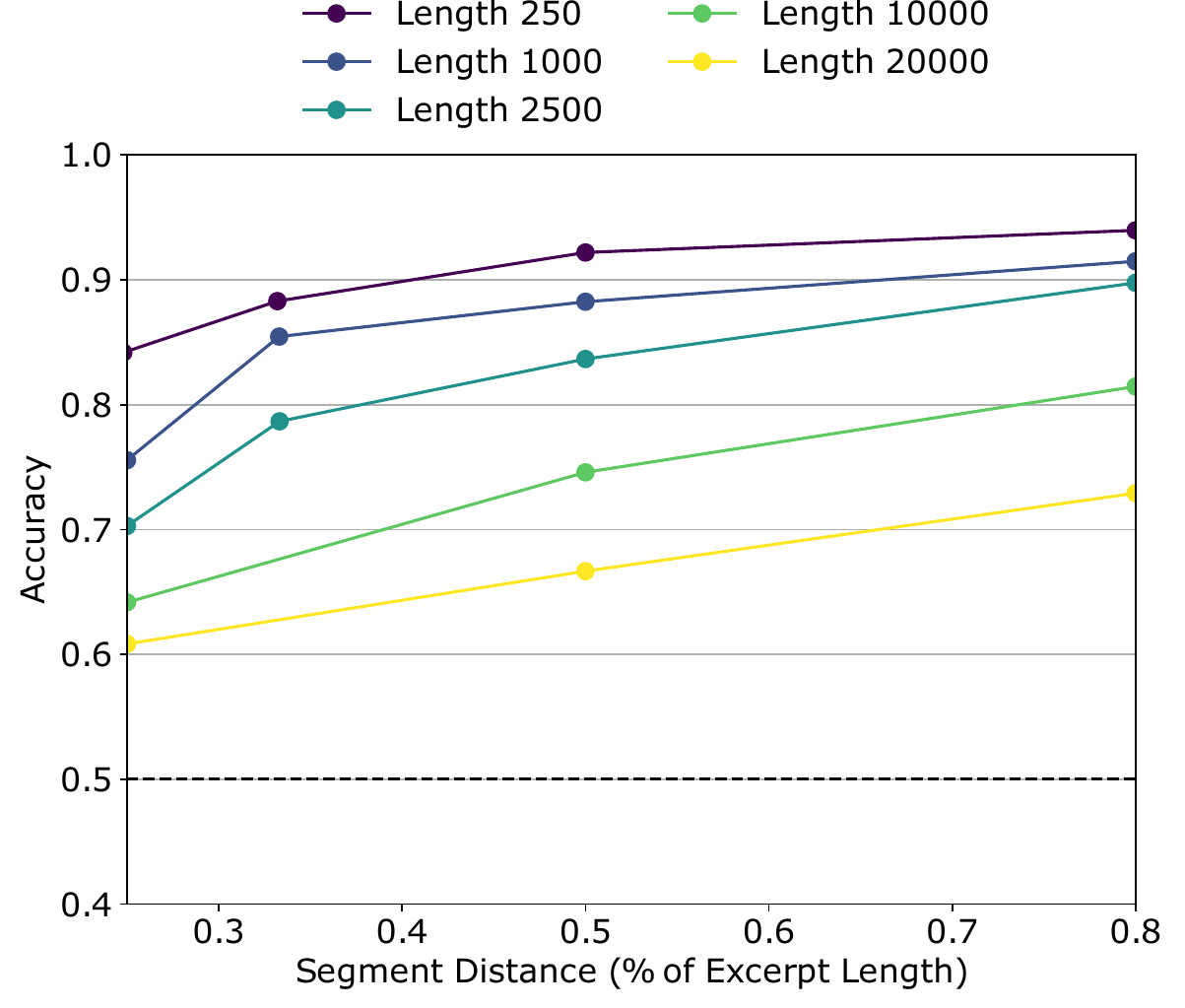}
        \caption{By segment distances (avg. over models)}
        \label{fig:distance_bins}
    \end{subfigure}
    \caption{Factors affecting SORT performance based on in-context memory. (a) SORT accuracy by excerpt length. (b) Average over SORT performance of different models across segment distances for different excerpt lengths.}
    \label{fig:combined}
\end{figure}

\textbf{In-context memory performance increases with greater distance between segments.} We further evaluate the effect of another factor which may influence the model performance--the distance between the text segments in the excerpt. Figure \ref{fig:distance_bins} shows an increasing trend in accuracy as the distance between segments increases. This improvement in accuracy, which we also observed in our human experiment (Fig. \ref{fig:humans-LTM}), is consistent across excerpt lengths and is observed across all models (see Appendix \ref{appendix: stm_full_results}).

\textbf{In-context memory performance decreases with increasing excerpt length.} Average performance on longer excerpts (Table \ref{tab:stm_results}, SORT-extend) is substantially lower than in the standard context lengths, despite the presence of longer segment distances.
For increasing excerpt lengths, we see a consistently monotonic decrease in average accuracy (Figures \ref{fig:excerpt_length} and \ref{fig:combined}). This is consistent with previous findings on length generalization in LLMs \citep{liu2024lost,levy2024same, hsieh2024ruler}.

\textbf{Additional analyses.} Further analyses are presented in Appendix \ref{appendix: stm_full_results}. Like humans, models handle longer segments (50 words) slightly more effectively than shorter segments (20 words), with an improvement of up to $4\%$. We found no significant differences across books from different domains (Table \ref{tab:accuracy_diff_results_part1}-\ref{tab:accuracy_diff_results_part2}).

\begin{table}[t]
    \centering
    \captionsetup{skip=1\baselineskip} 
    \caption{Mean of in-context memory performance with 95\% bootstrapped confidence interval. SORT-extend shows performance with excerpts of lengths 10000 and 20000 words, which exceeds most models' context lengths.}
    \begin{tabular}{l S[table-format=6.0] S[table-format=2.0]  S[table-format=2.0] S[table-format=2.0] S[table-format=2.0] S[table-format=2.0]}
        \toprule
        \textbf{Model name} & \textbf{Parameters} & \textbf{Max context} & \textbf{SORT} &\textbf{SORT-extend}\\
        \midrule
        Llama3-70b-inst & {70b}  & {8k} & {$0.92\pm0.020$} & {/}\\
        Llama3-8b-inst & {8b} & {8k} & {$0.93\pm0.007$} & /\\
        Mixtral-8x22b-inst & {8x22b} & {64k} & {$0.95\pm0.020$} & {$0.79\pm0.038$}\\
        Mixtral-8x7b-DPO-inst & {8x7b} & {32k} & {$0.89\pm0.030$} & {$0.56\pm0.058$}\\
        Llama2-70b-inst & {70b} & {8k}& {$0.77\pm0.040$}& /\\
        Gemma-1.1-7b-inst & {7b} & {8k} & {$0.85\pm0.010$} & /\\
        Mistral-v0.2-7b-inst & {7b}& {32k} & {$0.85\pm0.032$} & {$0.65\pm0.045$}\\
        Mistral-v0.1-7b-inst & {7b}& {8k} & {$0.77\pm0.013$} & / \\
        Llama2-7b-inst & {7b} & {4k} & {$0.56\pm0.014$} & /\\
        GPT-3.5-turbo & {unknown} & {16k} & {$0.86\pm0.010$} & {/}\\
        \bottomrule
    \end{tabular}
    
    \label{tab:stm_results}
\end{table}

\subsection{Parametric Memory via Finetuning}

\textbf{Full parameter fine-tuning on books with a language modeling objective did not improve SORT performance.} For Llama3-8b-Instruct and Mistral-7b-v0.2-Instruct, we do not observe any difference in performance on SORT after memory is inserted via fine-tuning on large chunks of book-text. A pairwise statistical analysis across epochs of fine-tuning, relative to two baselines that either exclude the books from the fine-tuning dataset or instead include only summaries of the books, shows no substantial improvement (see Appendix \ref{appendix:finetuning}). 

\subsection{Retrieval Augmented Memory}

\begin{figure}[t]
  \begin{subfigure}[t]{0.495\textwidth}
    \includegraphics[width=\textwidth]{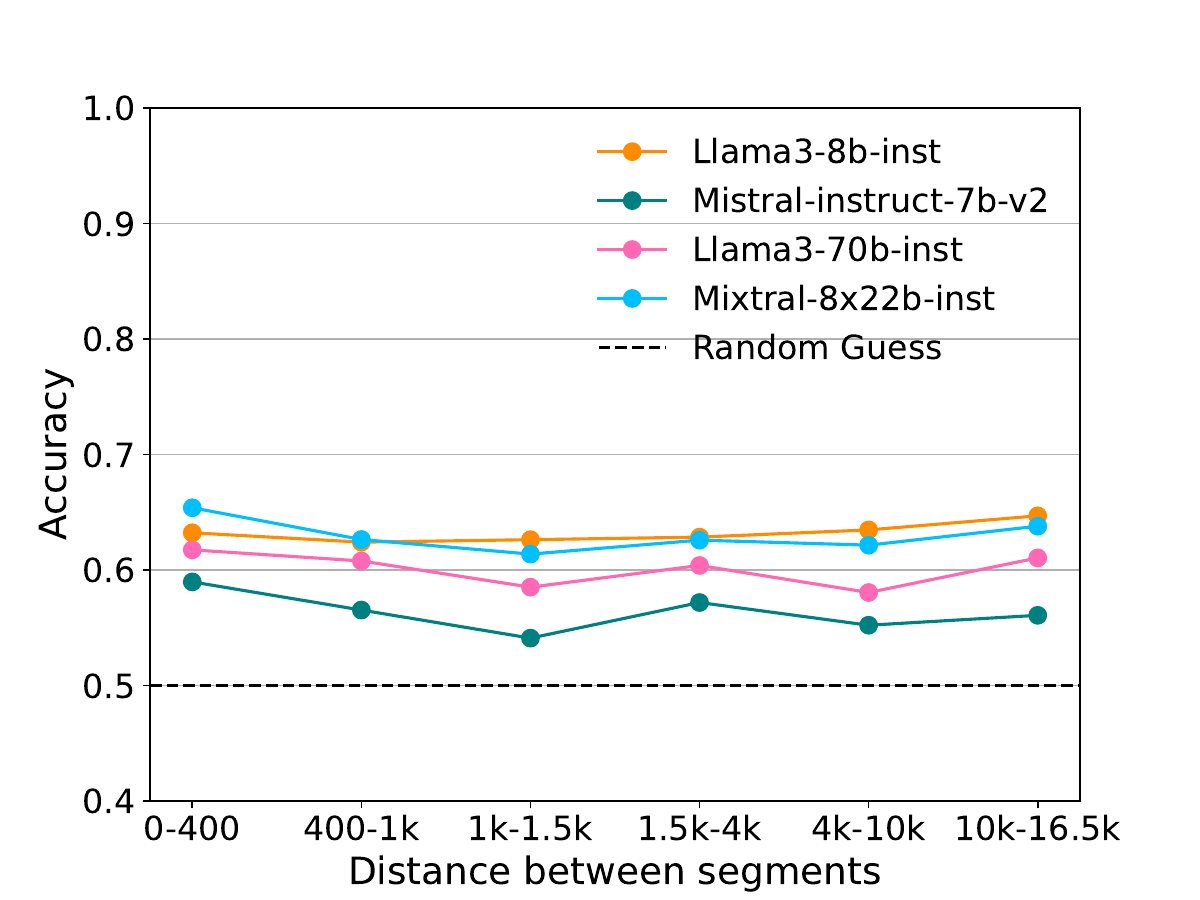}
        \caption{Vanilla RAG}
        \label{fig:rag-combine}
    \end{subfigure}
     \begin{subfigure}[t]{0.495\textwidth}
    \includegraphics[width=\textwidth]{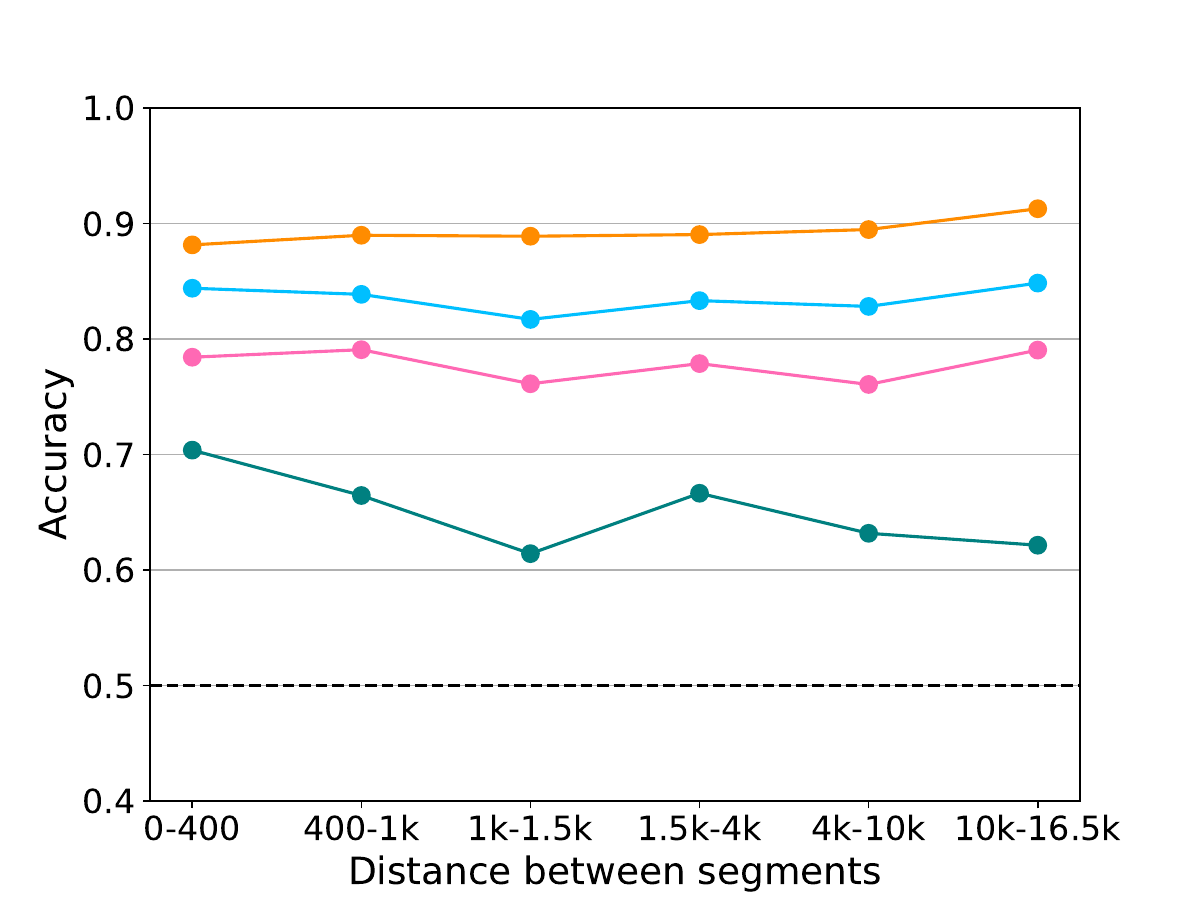}
        \caption{RAG with oracle retriever and order preservation} 
        \label{fig:rag-ordered}
    \end{subfigure}
    \caption{SORT performance based on RAG memory. (a) Accuracy with vanilla RAG memory. (b) Accuracy with RAG memory for those samples where the correct passages of text are retrieved and presented in the order in which they appeared in the books.}
    \label{fig:rag}
\end{figure}

\textbf{RAG based memory leads to worse performance than in-context memory.} RAG performance is between 55\% and 67\% for all distances between segments and tested models
(Figure \ref{fig:rag-combine}), which is substantially lower than the in-context memory performance. This difference in performance follows from the fact that standard forms of RAG do not necessarily preserve the order of retrieved passages, whereas the excerpt provided for in-context memory does have the passages in the correct order (and additionally contains the text that connects the passages, which may help in making the order judgment). When the relevant passages are retrieved and presented in the correct order, RAG performance improves substantially. Interestingly, we find that Llama3-8b-Instruct model outperforms the much larger Mixtral-8x22b-Instruct and Llama3-70b-Instruct on SORT with an accuracy around $90\%$ across all distances between segments (Figure \ref{fig:rag-ordered}).

\section{Discussion}
\label{sec:discuss}

We provide a new evaluation task, SORT, for assessing episodic memory in large language models, that can be used with any text data and without the need for annotation. We created Book-SORT, a dataset for SORT based on books that were recently added to the public domain and we validated that book-specific memory is indeed needed to achieve high performance on Book-SORT. We evaluated three different ways to include memory of specific texts in a model to assess whether they support a key function of episodic memory. Below, we discuss our results for these methods in relation to episodic memory in humans.

\textbf{Is in-context memory a form of episodic memory?} Several links have been drawn between in-context memory in transformers and models of episodic memory in humans \citep{jian2024linkingincontextlearningtransformers,whittington2022relating, Whittington2023.11.05.565662, ellwood2024}, and our results, which show that in-context memory supports sequence order recall, could be interpreted as further evidence for in-context memory acting as episodic memory in LLMs.
However, our results also show that in-context sequence order recall performance degrades with increasing context length, which would not be the case with episodic memory.
This discrepancy stems from a key difference between in-context memory in models and episodic memory in humans and animals, which is that in-context memory in LLMs can directly attend to all tokens in the context window, whereas the episodic memory system in humans and animals stores past experiences in synaptic form, and requires an additional retrieval step before episodic memory content can be attended to. The reliance on synaptic storage and retrieval is what enables the episodic memory system in humans and animals to make use of a sequence-length invariant mechanism with a fixed computational cost to remember past experiences over a lifetime. This sequence-length invariant property of the episodic memory system in humans and animals allows it to generalize to arbitrarily long sequences, while attention over all tokens in a growing sequence eventually leads to generalization failure for in-context memory and, at the same time, comes with a sharply increasing computational cost.
Based on these considerations, we believe that, although both the episodic memory system in animals and in-context memory in transformer models perform a kind of similarity-based lookup of past experiences, in-context memory's access to activations is more analogous to working memory in humans \citep{OReillyMunakataFrankEtAl24}, but with a capacity that vastly exceeds human working memory.

\textbf{Is parametric memory in transformers a form of episodic memory?} High performance on benchmarks including MMLU suggests that parametric memory in LLMs learned via a language modeling objective can support semantic forms of memory (e.g. when recalling knowledge to answer factual questions). Our evaluation on SORT showing close to chance performance after finetuning suggests that current forms of parametric memory do not support functions similar to those of episodic memory. This suggests that different learning methods and architectures (e.g. with a separate memory system) may be needed for functioning parametric forms of episodic memory.

\textbf{Is retrieval augmented memory a form of episodic memory?} Since it avoids the problems of context-length generalization and increasing computational costs observed for in-context memory, Retrieval Augmented Generation presents a potentially strong way to include memory of episodes via a retrieval process and subsequent in-context presentation. However, our results suggest that there is a lot of room for improvement over the performance of vanilla RAG. The weak performance of vanilla RAG on SORT arises from the fact that it is {\em decontextualized}--all that it retrieves is independent parts of the text. By contrast, current theories of episodic memory posit that episodic memory contents are bound to a drifting temporal context; later, when some content is retrieved, the temporal context associated with that content is also retrieved \citep{HOWARD2002269, Polyn2009}. One aspect of retrieved temporal context--absent from vanilla RAG--is required for sequence order recall. Order-preserving (OP) variants of RAG \citep{yu2024defenserageralongcontext} can increase performance on SORT, as suggested by our results shown in Figure \ref{fig:rag-ordered}. However, OP-RAG maintains contextual information only about the sequential order, and it does not bind any other temporal context to the independently retrieved passages. The core difference between current RAG systems and episodic memory remains: they do not present a method to bind temporal context to the content of memories.

\textbf{Limitations.} Current high performing LLMs do not disclose their training data, which means that care needs to be taken in selecting suitable data to include in a SORT dataset. To minimize the probability that models have been trained on books used for our SORT evaluation, we curated Book-SORT based on books that were not publicly available when models were trained. However we cannot rule out the possibility that the books in this set were used in training of a model, which (if true) would require us to interpret results as indicating the effectiveness of additional rather than initial memory-insertion. Furthermore the reliance on instruction-following can limit the applicability to both non-instruction-tuned models and models that have poor instruction-following ability. 
Lastly, we provided two examples of more long-term memory-insertion via fine-tuning and Retrieval Augmented Generation for two models, Llama3-8b-Instruct and Mistral-7b-v0.2-Instruct, and leave more extensive studies on how to induce episodic memories without relying on complete in-context presentation to future work.

\textbf{Future work.} Improving long-term memory in LLMs is an emerging area of research \citep{ltm-liu2023think,ltm-borgeaud2022improving,ltm-fournier2023practical,ltm-phang2023investigating,ltm-wang2024augmenting,ltm-zhong2022training,ltm-zhong2024memorybank}, and SORT can be used to assess improvement in an crucial aspect of an important form of memory in new models. Specifically, improving episodic memory in models may improve models' continual learning, performance on tasks at long contexts such as extended chat exchanges with a user, and source attribution via knowledge of where and when a memory was acquired. Recent efforts have highlighted the potential of augmenting LLMs with additional episodic memory mechanisms \citep{fountas2024humanlikeepisodicmemoryinfinite, das2024larimarlargelanguagemodels}, and we expect that SORT can be used to evaluate these classes of models, once such a model with a sufficiently strong instruction-following ability is released. Another possibility is to identify new and better methods to insert episodic memory of texts into existing models. Additionally, SORT can be extended to other types of inputs, such as audio and video, which can be used to evaluate episodic memory in multimodal models in the future.

\textbf{Conclusion.} The ability of LLMs to retain and retrieve long-term knowledge is crucial for their continued integration in many applications. Therefore, a more comprehensive and systematic evaluation of these abilities is needed. We believe that the new evaluation framework SORT offers a promising path for future research aimed at better understanding and improving these capabilities in foundation models.

\textbf{Ethics Statement.} To avoid ethical issues concerning copyright, we based Book-SORT on books that were recently added to the public domain. Our human experiment with 155 participants was approved by the IRB at Anonymized University and participants were compensated.

\textbf{Reproducibility Statement.} We will publicly release the Book-SORT dataset as well as all code to generate new SORT datasets and evaluate models on SORT. For open models, evaluation on Book-SORT is deterministic due to greedy sampling and the use of an answer prefix.

\textbf{Acknowledgements}
Funded in part by the Deutsche Forschungsgemeinschaft (DFG, German Research Foundation) -- GRK 2853/1 “Neuroexplicit Models of Language, Vision, and Action” - project number 471607914.

\bibliographystyle{arxiv/iclr2025_conference}
\bibliography{main}

\newpage

\appendix

\section{Additional details on \datasetname data set}

\textbf{Preprocessing book text.} We wrote custom Python code to only retain the book text that formed a continuous narrative. We stripped the front and back matter of the book, and extracted chapter titles if they existed. 8 of the 9 books contained individual section or chapter breaks. For these 8 books, we parsed the text corresponding to each chapter. Chapter titles or section headings (e.g. `VI' to indicate section six) were removed, and all remaining text was concatenated. This string was split into words (assuming simple whitespace separators with Python \texttt{string.split()}) to produce a final text array for each book. This text array was sampled for the \datasetname dataset.

\subsection{Book selection}

We provide details about the $9$ books in Book-SORT in Table \ref{tab:books}.

\label{appendix:books}

\vspace{1\baselineskip}
\begin{table}[h]
\centering
\captionsetup{skip=1\baselineskip} 
\caption{Project Gutenberg metadata on \datasetname books.}
\label{tab:books}
\resizebox{\textwidth}{!}{
    \begin{tabular}{@{}clp{3cm}lcccp{5cm}@{}}
    \toprule
    ID     & Title                              & Author                              & Word count & Release   & Pub  & LoCC*  & Subjects                                                                                                                                    \\ \midrule
    69087  & The Murder of Roger Ackroyd        & Christie, Agatha                    & 69,720      & 10/2/2022 & 1926 & PR   & Detective and mystery stories; Fiction: Private investigators - England, Murder - Investigation, Belgians - England \\
    72578 & Tom Swift and His Talking Pictures & Appleton, Victor                    & 43,853      & 1/1/2024  & 1928 & PZ   & Adventure stories; Motion pictures                                                                                                          \\
    72600  & The Trumpeter of Krakow            & Kelly, Eric Philbrook     & 59,081      & 1/2/2024  & 1928 & PZ   & Juvenile fiction: Middle Ages, Poland - History - Casimir IV, 1447-1492                                                             \\
    72869  & Meet the Tiger                     & Charteris, Leslie                   & 79,946      & 2/4/2024  & 1928 & PR   & Fiction: Private investigators - England; Detective and mystery stories                                                                 \\
    72958  & Hunting for Hidden Gold            & Dixon, Franklin W.                  & 42,354      & 2/14/2024 & 1928 & PZ   & Juvenile fiction: Brothers, Gold mines and mining, Montana, Robbers and outlaws; Mystery and detective stories                              \\
    72963  & The Nature of the Physical World   & Eddington, Arthur Stanley, Sir      & 104,530     & 2/15/2024 & 1928 & Q    & Physics - Philosophy; Science - Philosophy                                                                                          \\
    72972  & Money for Nothing                  & Wodehouse, P.G. (Pelham Grenville) & 82,331      & 2/16/2024 & 1928 & PR   & Humorous stories; Fiction: Swindlers and swindling, Greed                                                                                   \\
    73017  & Pomona; or, the Future of English  & De Selincourt, Basil                & 9,273       & 2/22/2024 & 1928 & PE   & English language                                                                                                                            \\
    73042  & The Well of Loneliness             & Hall, Radclyffe                     & 163,217     & 2/26/2024 & 1928 & PR   & Fiction: Lesbians - England - Social conditions                                                                                   \\ \bottomrule
    \end{tabular}
}
\end{table}
{\footnotesize *LoCC = Library of Congress classification.}

\subsection{Between-segment distances}

The segment distance $L_S$ for Book-SORT is sampled from one of four distance bins. The right edge of each bin is given in Table \ref{tab:segdist}. Distance is computed between the beginning of the first segment and the beginning of the second segment. The minimum distance $L_S$ therefore produces adjacent, non-overlapping segments.

\begin{table}[h]
\centering\captionsetup{skip=1\baselineskip} 
\caption{Right edge of each distance bin used to create samples for Book-SORT.}
\label{tab:segdist}
\begin{tabular}{|l|l|l|l|l|l|}
\hline
               & Minimum & Bin0      & Bin1      & Bin2      & Bin3        \\ \hline
$L_E \leq 2,500$ & $L_S$   & $L_E / 4$ & $L_E / 3$ & $L_E / 2$ & $L_E / 0.8$ \\ \hline
$L_E \geq 10,000$ & $L_S$   & 1000      & $L_E / 4$ & $L_E / 2$ & $L_E / 0.8$ \\ \hline
\end{tabular}
\end{table}

\subsection{Human study details}
\label{appendix:human_exp}

\paragraph{Participant compensation.} Participants were compensated via a lottery system with a chance to win a gift card to a popular book store. The expected value of the compensation came out to $\$12$ per hour.

\paragraph{Study design.} Each participant completed an online survey. First, the participant consented to the study, read a brief set of instructions, and completed a brief survey, including a question regarding when the participant finished reading the book. The complete set of survey questions is listed below. Each participant was then asked to answer "Which segment occurred first in the book?" for $10$ randomly chosen text segment pairs from a total set of $540$ unique segment pairs sampled from the whole book. We chose to present a sample number of trials to each participant to minimize interference effects from repeated memory retrieval \citep{kliegl2021mechanisms}. The presentation order of the text segments was randomized across participants. In the end, each participant was asked $4$ simple questions about the book plot to verify that the participant had indeed read the book. Each participant was only allowed to participate in the study once.

\paragraph{Demographics questions.} The human participants were asked the following set of demographics questions before beginning the experiment:

\begin{enumerate}
    \item I have finished the book The Murder of Roger Ackroyd [Options: True/False]
    \item On what date did you finish the book? [Calendar question type]
    \item Did you read or listen to the book? [Options: Read/Listen]
    \item Was this your first time reading / listening to the book? [Options: Yes / No]
    \item What is your age? [Options: 18-25, 25-35, 35-45, 45-55, 55-65, 65+]
    \item What gender do you identify with? [Options: Female/Male/Other]
    \item What is your experience with the English language? [Options: Native / Fluent / Advanced / Intermediate / Beginner] 
    \item How many books did you read or listen to in the past year? [Options: 1-2 / 3-5 / 6-10 / 10+]
\end{enumerate}

We use the responses above to determine the number of days that have passed since finishing the book, and make this information available in the human dataset together with the responses.

\paragraph{Inclusion criteria.} We include data from participants who answered at least $3$ of $4$ plot questions correctly, and finished reading the book within $30$ days of participating in the study. These inclusion criteria result in $155$ participants.

\section{Model and prompting details}
\subsection{Model details}
\label{appendix:models}
We listed all models we used in this paper and their download links from HuggingFace in Table \ref{table:all_models_name}. For the OpenAI models, we used the gpt-3.5-turbo-0125 version of GPT-3.5, and gpt-4-turbo-2024-04-09 for GPT-4. Models were selected to cover a broad range of performance on more semantic/knowledge-based tasks such as those included in MMLU.

\vspace{1\baselineskip} 
\begin{table}[h]
\centering
\captionsetup{skip=1\baselineskip} 
\caption{Model Details}
\begin{tabular}{c c c}
\toprule
\textbf{Name in HuggingFace} & \textbf{Name in Paper} & \textbf{MMLU score}\\ \hline
\href{https://huggingface.co/meta-llama/Meta-Llama-3-70B-Instruct}{Llama-3-70B-Instruct} & Llama3-70b-inst & 80.06\\ 
\href{https://huggingface.co/meta-llama/Meta-Llama-3-8B-Instruct}{Llama-3-8B-Instruct} & Llama3-8b-inst & 66.60\\ 
\href{https://huggingface.co/mistralai/Mixtral-8x22B-Instruct-v0.1}{Mixtral-8x22B-Instruct-v0.1} & Mixtral-8x22b-inst & 77.77 \\
\href{https://huggingface.co/NousResearch/Nous-Hermes-2-Mixtral-8x7B-DPO}{Nous-Hermes-2-Mixtral-8x7B-DPO} & Mixtral-8x7b-DPO-inst & 72.28\\
\href{https://huggingface.co/mistralai/Mistral-7B-Instruct-v0.1}{Mistral-7B-Instruct-v0.1} & Mistral-v1-7b-inst & 60.10 \\
\href{https://huggingface.co/mistralai/Mistral-7B-Instruct-v0.2}{Mistral-7B-Instruct-v0.2} & Mistral-v2-7b-inst &60.07\\ 
\href{https://huggingface.co/meta-llama/Llama-2-70b-chat}{Llama-2-70b-chat} & Llama2-70b-inst &68.90\\ 
\href{https://huggingface.co/meta-llama/Llama-2-7b-chat}{Llama-2-7b-chat} & Llama2-7b-inst &	45.30 \\ 
\href{https://huggingface.co/google/gemma-1.1-7b-it}{gemma-1.1-7b-inst} & Gemma-1.1-7b-inst &	64.30\\ 
\toprule
\end{tabular}
\label{table:all_models_name}
\end{table}
\vspace{1\baselineskip} 

\subsection{Prompting}
\label{appendix:prompts}

For our experiments with Book-SORT, we created a total of 12 prompts that are composed of two parts. The prompts differ in how they phrase the tasks. The first part contains instructions to read the text excerpt from the book as well as a placeholder for the actual excerpt. The second part of the prompt contains the description of SORT, including a mention of the book or document title as well as two segments from that document. We found that current open LLMs fail at the task even with in-context access to the text, if they are asked to tell which segment appeared second or last. For this reason, we ran all experiments with the placeholder <position> set to "first". All of these prompts were preceded by the same generic system prompt: \textit{"You are a helpful, respectful and honest assistant."}

\begin{longtable}{r>{\raggedright\arraybackslash}p{0.42\textwidth}>{\raggedright\arraybackslash}p{0.42\textwidth}}
\caption{Selection of 12 prompts used for prompt validation} \\
\hline
\textbf{No.} & \textbf{Reading instruction} & \textbf{SORT instruction} \\
\hline
\endfirsthead
\caption{Selection of 13 prompts used for prompt validation} \\
\hline
\textbf{No.} & \textbf{Reading instruction} & \textbf{SORT instruction} \\
\hline
\endhead

\hline
\endfoot

1 & "Please take some time to thoroughly read and comprehend this extract from the book <booktitle>. The passage is as follows: <excerpt>" & "You will be shown pairs of text fragments from <booktitle>. Please select which of two fragments appeared <position> in the book. You will be shown 10 such pairs. <segments> Which fragment appeared <position> in the book, <label\_0> or <label\_1>?"
 \\
\hline
2 & "I need you to thoroughly read and comprehend this extract from the book <booktitle>. The passage is as follows: <excerpt>"
 & "In this exercise, your objective is to identify the text segment, either <label\_0> or <label\_1>, that appeared <position> in <booktitle>. Please read the segments carefully to determine their order of appearance in <booktitle> and respond with either <label\_0> or <label\_1>: <segments> Which of these, <label\_0> or <label\_1>, was <position> in <booktitle>?"
 \\
\hline
3 & "I need you to thoroughly read and comprehend this extract from the book <booktitle>. The passage is as follows: <excerpt>"
 & "Your task is to recall which text segment, either <label\_0> or <label\_1>, appeared <position> in the book <booktitle>. Please read the segments carefully to remember in which order they appeared in <booktitle> and respond with either <label\_0> or <label\_1>: <segments> Which of these, <label\_0> or <label\_1>, was <position> in the book <booktitle>?"
 \\
\hline
4 & "I need you to thoroughly read and comprehend this extract from the book <booktitle>. The passage is as follows: <excerpt>"
 & "You will be shown two text segments, labeled as <label\_0> and <label\_1>. Please recall in which order they appeared in the book <booktitle> and tell me which one came <position>. Please read the segments carefully: <segments> Which of these two parts of the book, <label\_0> or <label\_1>, came <position> in the book <booktitle>?"
 \\
\hline
5 & "I need you to thoroughly read and comprehend this extract from the book <booktitle>. The passage is as follows: <excerpt>"
 & "I will show you two short parts from a book, labeled as <label\_0> or <label\_1>. Your task is to tell me which of them appeared <position> in the book <booktitle>. Please read both segments carefully and try to remember where in the book they come from: <segments> Which of these, <label\_0> or <label\_1>, appeared <position> in the book <booktitle>?"
 \\
\hline
6 & "I need you to thoroughly read and comprehend this extract from the book <booktitle>. The passage is as follows: <excerpt>"
 & "This is your task: Given two segments from a book, labeled as <label\_0> and <label\_1>, please tell me which of them appeared <position> in <booktitle>. Read both segments carefully and try to remember where in <booktitle> they appeared: <segments> Which of these, <label\_0> or <label\_1>, comes <position> in the book <booktitle>?"
 \\
\hline
7 & "Please carefully read this excerpt from the book <booktitle>. This is the relevant passage: <excerpt>"
 & "You will be shown pairs of text fragments from <booktitle>. Please select which of two fragments appeared <position> in the book. You will be shown 10 such pairs. <segments> Which fragment appeared <position> in the book, <label\_0> or <label\_1>?"
 \\
\hline
8 & "Please carefully read this excerpt from the book <booktitle>. This is the relevant passage: <excerpt>"
 & "In this exercise, your objective is to identify the text segment, either <label\_0> or <label\_1>, that appeared <position> in <booktitle>. Please read the segments carefully to determine their order of appearance in <booktitle> and respond with either <label\_0> or <label\_1>: <segments> Which of these, <label\_0> or <label\_1>, was <position> in <booktitle>?"
 \\
\hline
9 & "Please carefully read this excerpt from the book <booktitle>. This is the relevant passage: <excerpt>"
 & "Your task is to recall which text segment, either <label\_0> or <label\_1>, appeared <position> in the book <booktitle>. Please read the segments carefully to remember in which order they appeared in <booktitle> and respond with either <label\_0> or <label\_1>: <segments> Which of these, <label\_0> or <label\_1>, was <position> in the book <booktitle>?"
 \\
\hline
10 & "Please carefully read this excerpt from the book <booktitle>. This is the relevant passage: <excerpt>"
 & "You will be shown two text segments, labeled as <label\_0> and <label\_1>. Please recall in which order they appeared in the book <booktitle> and tell me which one came <position>. Please read the segments carefully: <segments> Which of these two parts of the book, <label\_0> or <label\_1>, came <position> in the book <booktitle>?"
 \\
\hline
11 & "Please carefully read this excerpt from the book <booktitle>. This is the relevant passage: <excerpt>"
 & "I will show you two short parts from a book, labeled as <label\_0> and <label\_1>. Your task is to tell me which of them appeared <position> in the book <booktitle>. Please read both segments carefully and try to remember where in the book they come from: <segments> Which of these, <label\_0> or <label\_1>, appeared <position> in the book <booktitle>?"
 \\
\hline
12 & "Please carefully read this excerpt from the book <booktitle>. This is the relevant passage: <excerpt>"
 & "This is your task: Given two segments from a book, labeled as <label\_0> and <label\_1>, please tell me which of them appeared <position> in <booktitle>. Read both segments carefully and try to remember where in <booktitle> they appeared: <segments> Which of these, <label\_0> or <label\_1>, comes <position> in the book <booktitle>?"
 \\
\hline

\label{appendix:prompt-list}
\end{longtable}

\subsection{Per-model results on prompt selection sweep}
\label{appendix:prompts-results}

To identify the prompts that work best for each model, we take 400 segment-pair samples that we excluded from the main evaluation and evaluate models' in-context memory with all prompts shown in Table \ref{appendix:prompt-list}. To select the best prompt we considered both the proportion of A and B responses, which should be around $0.5$, and the accuracy. 
We report the best selected prompts in Table \ref{appendix:prompt-validation-results} with numbers referring to the prompts presented in Table \ref{appendix:prompt-list}.
\begin{table}[h]
    \centering
    \caption{Selected prompts for each model.}
    \label{appendix:prompt-validation-results}
    \begin{tabular}{lc}
        \toprule
        \textbf{Model Name} & \textbf{Best Prompt} \\
        \midrule
        Llama3-70b-inst & 4 \\
        Llama3-8b-inst & 3 \\
        Mixtral-8x22b-inst & 4 \\
        Llama2-70b-inst & 7 \\
        Gemma-1.1-7b-inst & 8 \\
        Mistral-v0.2-7b-inst & 3 \\
        Mistral-v0.1-7b-inst & 2 \\
        Llama2-7b-inst & 10 \\
        GPT-3.5-turbo  & 7 \\ 
        GPT-4 & 7 \\
        \bottomrule
    \end{tabular}
\end{table}

\subsection{RAG prompt selection}
\label{appendix:prompts-rag}
There were two different prompts to select for the retrieval-augmented generation experiments: the retrieval prompt (i.e. the search query), and the LLM prompt.

\subsubsection{Retrieval prompt (search query)}
The goal of retrieval in our RAG experiments is to find the text passages that will provide the most information about the segments for the sequence ordering task. After we created the vector database of all the text passages from Book-SORT, we formulated several different search queries (Table \ref{appendix:rag-search-queries}). We then ran retrieval using a validation subset of \datasetname (50-word segments, 250-word excerpts from all books). The retrieval used the same database and text embedding model as described in the RAG portion of Section \ref{subsec:memory_methods}. The best search query was simple and only consisted of the segment text (query 8, Table \ref{appendix:rag-search-queries}). This search query is used for all RAG experiments.

\begin{table}[h]
    \centering
    \caption{The search queries for the RAG experiment and their average retrieval recall@10 on a validation subset of \datasetname(250 word excerpts, 50 word segments).}
    \label{appendix:rag-search-queries}
    \begin{tabular}{r p{4in} >{\centering\arraybackslash}p{0.75in}}
        \toprule
        \textbf{No.} & \textbf{Search Query Text} & \textbf{Recall@10} \\ \midrule
        0 & "Please determine the order in which the following text segments appeared in \textless{}booktitle\textgreater{}: \textless{}segments\textgreater{}" & 0.728 \\
        1 & "We need to put text segments from \textless{}booktitle\textgreater{} in order. These are the segments: \textless{}segments\textgreater{}" & 0.817 \\
        2 & "Please find these text segments from \textless{}booktitle\textgreater{}: \textless{}segments\textgreater{}" & 0.869 \\
        3 & "Please find these text segments from \textless{}booktitle\textgreater{} to provide context for the next task: \textless{}segments\textgreater{}" & 0.875 \\
        4 & "Which text chunks from \textless{}booktitle\textgreater{} contain the following segments? \textless{}segments\textgreater{}" & 0.802 \\
        5 & "Which text excerpts from \textless{}booktitle\textgreater{} contain the following segments? \textless{}segments\textgreater{}" & 0.799 \\
        6 & "Which text chunks from \textless{}booktitle\textgreater{} overlap with these text segments: \textless{}segments\textgreater{}" & 0.782 \\
        7 & "\textless{}booktitle\textgreater{} contains this text: \textless{}segments\textgreater{}" & 0.865 \\
        8 & "\textless{}segments\textgreater{}" & 0.906 \\
        9 & "\textless{}booktitle\textgreater{} \textless{}segments\textgreater{}" & 0.858 \\ \bottomrule
    \end{tabular}
\end{table}

\subsubsection{RAG LLM prompts}
We followed a procedure similar to the one outlined in Section \ref{appendix:prompts}. We created a total of 10 modifications to the reading instructions from Table \ref{appendix:prompt-list}.

\begin{table}[h]
    \centering
    \caption{RAG prompt modifications.}
    \label{appendix:rag-prompt-list}
    \begin{tabular}{rp{4.8in}}
    \toprule
    \textbf{No.} & \textbf{RAG Reading Instruction} \\ \midrule
    0 & "Here are some relevant excerpts from the book \textless{}booktitle\textgreater{}: \textless{}context\textgreater{}" \\
    1 & "The following excerpts from the book \textless{}booktitle\textgreater may be helpful context for the task. Context: \textless{}context\textgreater{}" \\
    2 & "Context: \textless{}context\textgreater{}" \\
    3 & "Searching a book database found these relevant text snippets: \textless{}context\textgreater{}" \\
    4 & "The following search results may be useful context: \textless{}context\textgreater{}" \\
    5 & "I will show you some relevant text found by searching a database of books: \textless{}context\textgreater{}" \\
    6 & "Please read some text deemed relevant for the task before performing the task. Relevant text: \textless{}context\textgreater{}" \\
    7 & "Please read these search results carefully to help you perform the task. Search results: \textless{}context\textgreater{}" \\
    8 & "Your objective may become easier with the use of these search results: \textless{}context\textgreater{}" \\
    9 & "This context may be helpful: \textless{}context\textgreater{}" \\ \bottomrule
    \end{tabular}
\end{table}

\subsubsection{Per-model results on RAG prompt selection}
For a given LLM, we modified the reading instruction of the best prompt from Table \ref{appendix:prompt-validation-results} with each of the 10 options in Table \ref{appendix:rag-prompt-list}. We then ran a sweep over the same 400 segment-pair samples detailed in Section \ref{appendix:prompts-results} and found the instruction that resulted in the highest performance on this held-out dataset.

\begin{table}[h]
    \centering
    \caption{Best RAG instruction prompts for each model.}
    \label{appendix:prompt-validation-results}
    \begin{tabular}{lc}
        \toprule
        \textbf{Model Name} & \textbf{Best RAG Instruction No.} \\
        \midrule
        Llama3-70b-inst & 7 \\
        Llama3-8b-inst & 7 \\
        Mixtral-8x22b-inst & 3 \\
        Mistral-v0.2-7b-inst & 6 \\
        \bottomrule
    \end{tabular}
\end{table}

\section{Additional details on \datasetname$ $ results}

\subsection{Memory-less baseline results}

\begin{figure*}[h]
    \begin{subfigure}[t]{1.0\textwidth}
        \centering
        \includegraphics[width=0.6\textwidth]{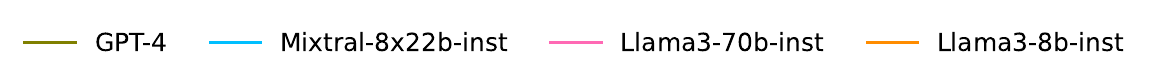}
    \end{subfigure}\\
    \begin{subfigure}[t]{0.5\textwidth}
        \centering
        \includegraphics[height=2in]{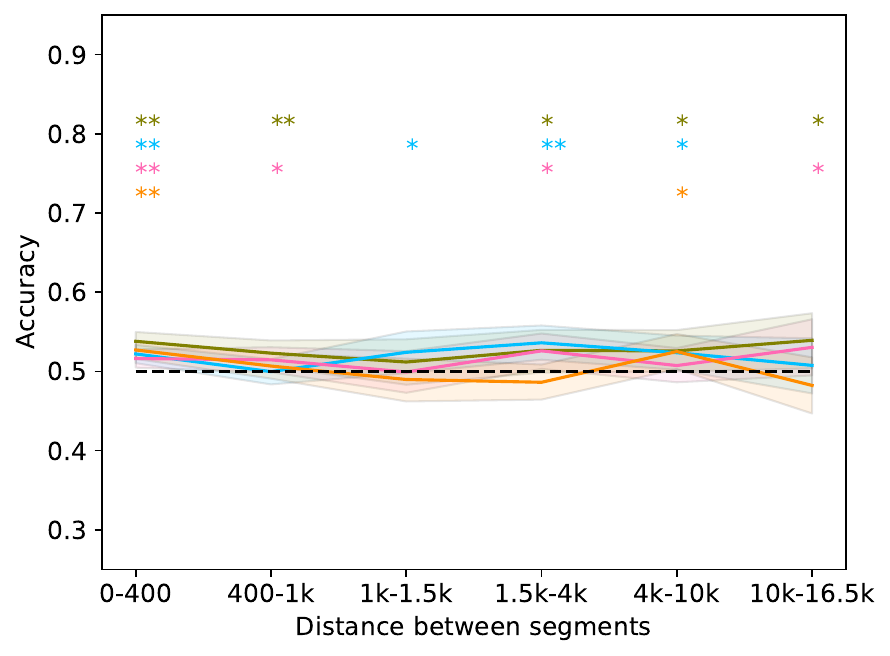}
        \caption{Segment length 20}
    \end{subfigure}%
    \begin{subfigure}[t]{0.5\textwidth}
        \centering
        \includegraphics[height=2in]{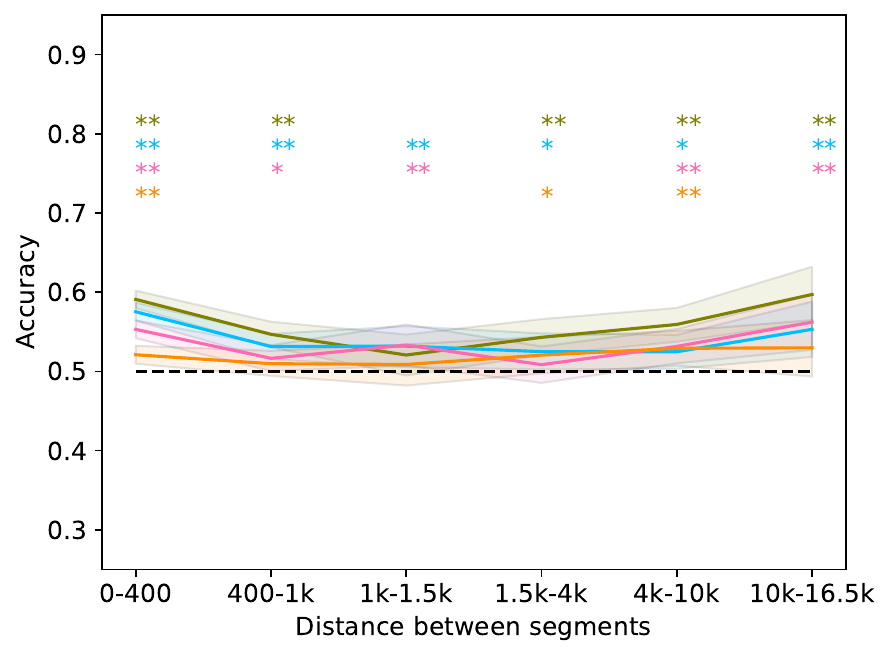}
        \caption{Segment length 50}
    \end{subfigure}
    \caption{Baseline SORT performance without memory of books in Book-SORT. Significant difference from chance is marked with asterisks ($^*$p-value$<$0.05,$^{**}$p-value$<$0.01).}
    \label{fig:LTM-distance}
\end{figure*}

Figure \ref{fig:LTM-distance} shows performance on Book-SORT without any memory-insertion of the books used in Book-SORT. We find that performance is higher in segment pairs that are very proximal or very distant in the book, indicating that it might be easier to sort these pairs based on temporal order reasoning. Performance without additional memory-insertion is generally low, showing that memory is needed for SORT.

\label{appendix:baseline}

\subsection{In-context memory full results}
\label{appendix: stm_full_results}
In this section, we provide a comprehensive overview of the in-context memory results across various models in Table \ref{tab:accuracy_diff_results_part1} and Table \ref{tab:accuracy_diff_results_part2}. The table below illustrates the accuracy of different models on multiple books at segment lengths of 20 and 50 words. We observe that, while models generally perform slightly better with longer segments (50 words) compared to shorter ones (20 words), the improvement is modest, averaging up to $4\%$.

\vspace{1\baselineskip} 
\begin{table}[t]
    \centering
    \captionsetup{skip=1\baselineskip}
    \caption{Accuracy and Difference of Various Models on Multiple Books at Excerpt Lengths of 20 and 50, with in-context memory (Part 1)}
    \begin{tabular}{l S[table-format=6.0] S[table-format=1.3] S[table-format=1.3] S[table-format=1.3] S[table-format=1.3] }
        \toprule
        \textbf{Model name} & \textbf{Book} & \textbf{SORT S20} & \textbf{SORT S50} &\textbf{SORT-Extend S20} & \textbf{SORT-Extend S50}  \\
        \midrule
        Llama3-8b-inst & {69087} & {0.89$\pm$0.03} & {0.92$\pm$0.03} & /&/ \\
        Llama3-8b-inst & {72578} & {0.91$\pm$0.02} & {0.93$\pm$0.02}& /&/ \\
        Llama3-8b-inst & {72600} & {0.92$\pm$0.03} & {0.94$\pm$0.02}& /&/ \\
        Llama3-8b-inst & {72869} & {0.92$\pm$0.03} & {0.94$\pm$0.02}& /&/ \\
        Llama3-8b-inst & {72958} & {0.92$\pm$0.02} & {0.94$\pm$0.02}& /&/ \\
        Llama3-8b-inst & {72963} & {0.92$\pm$0.03} & {0.94$\pm$0.02}& /&/ \\
        Llama3-8b-inst & {72972} & {0.92$\pm$0.03} & {0.94$\pm$0.02}& /&/ \\
        Llama3-8b-inst & {73017} & {0.91$\pm$0.03} & {0.94$\pm$0.02}& /&/ \\
        Llama3-8b-inst & {73042} & {0.92$\pm$0.03} & {0.94$\pm$0.02}& /&/ \\
        Llama2-70b-inst & {69087} & {0.74$\pm$0.12} & {0.90$\pm$0.08}& /&/ \\
        Llama2-70b-inst & {72578} & {0.75$\pm$0.12} & {0.90$\pm$0.09}& /&/ \\
        Llama2-70b-inst & {72600} & {0.71$\pm$0.13} & {0.91$\pm$0.09}& /&/ \\
        Llama2-70b-inst & {72869} & {0.71$\pm$0.13} & {0.91$\pm$0.09}& /&/ \\
        Llama2-70b-inst & {72958} & {0.71$\pm$0.13} & {0.90$\pm$0.09}& /&/ \\
        Llama2-70b-inst & {72963} & {0.72$\pm$0.13} & {0.89$\pm$0.10}& /&/ \\
        Llama2-70b-inst & {72972} & {0.70$\pm$0.13} & {0.88$\pm$0.10}& /&/ \\
        Llama2-70b-inst & {73017} & {0.70$\pm$0.13} & {0.87$\pm$0.10}& /&/ \\
        Llama2-70b-inst & {73042} & {0.71$\pm$0.13} & {0.88$\pm$0.10}& /&/ \\
        Llama2-7b-inst & {69087} & {0.56$\pm$0.05} & {0.56$\pm$0.05}& /&/ \\
        Llama2-7b-inst & {72578} & {0.57$\pm$0.05} & {0.55$\pm$0.05}& /&/ \\
        Llama2-7b-inst & {72600} & {0.57$\pm$0.05} & {0.56$\pm$0.04}& /&/ \\
        Llama2-7b-inst & {72869} & {0.57$\pm$0.05} & {0.56$\pm$0.04}& /&/ \\
        Llama2-7b-inst & {72958} & {0.57$\pm$0.05} & {0.56$\pm$0.04}& /&/ \\
        Llama2-7b-inst & {72963} & {0.57$\pm$0.05} & {0.57$\pm$0.05}& /&/ \\
        Llama2-7b-inst & {72972} & {0.57$\pm$0.05} & {0.56$\pm$0.05}& /&/ \\
        Llama2-7b-inst & {73017} & {0.57$\pm$0.05} & {0.56$\pm$0.05}& /&/ \\
        Llama2-7b-inst & {73042} & {0.57$\pm$0.05} & {0.56$\pm$0.05}& /&/ \\
       Llama3-70b-inst &{69087} & {0.90$\pm$0.08} & {0.92$\pm$0.09}& /&/ \\
        Llama3-70b-inst &{72578} & {0.92$\pm$0.08} & {0.92$\pm$0.09}& /&/ \\
        Llama3-70b-inst &{72600} & {0.92$\pm$0.08} & {0.93$\pm$0.09}& /&/ \\
        Llama3-70b-inst &{72869} & {0.93$\pm$0.07} & {0.93$\pm$0.08}& /&/ \\
        Llama3-70b-inst &{72958} & {0.93$\pm$0.07} & {0.94$\pm$0.08}& /&/ \\
        Llama3-70b-inst &{72963} & {0.92$\pm$0.08} & {0.93$\pm$0.09}& /&/ \\
        Llama3-70b-inst &{72972} & {0.91$\pm$0.08} & {0.93$\pm$0.09}& /&/ \\
        Llama3-70b-inst &{73017} & {0.92$\pm$0.08} & {0.94$\pm$0.09}& /&/ \\
        Llama3-70b-inst &{73042} & {0.91$\pm$0.09} & {0.94$\pm$0.08}& /&/ \\
        \bottomrule
    \end{tabular}
    \label{tab:accuracy_diff_results_part1}
\end{table}
\vspace{1\baselineskip} 

\newpage
\vspace{1\baselineskip} 
\begin{table}[t]
    \centering
    \captionsetup{skip=1\baselineskip}
    \caption{Accuracy and Difference of Various Models on Multiple Books at Excerpt Lengths of 20 and 50, with in-context memory (Part 2)}
    \begin{tabular}{l S[table-format=6.0] S[table-format=1.3] S[table-format=1.3] S[table-format=1.3] S[table-format=1.3] }
        \toprule
        \textbf{Model name} & \textbf{Book} & \textbf{SORT S20} & \textbf{SORT} &\textbf{SORT-Extend S20} & \textbf{SORT-Extend S50}  \\
        \midrule
        Mixtral-8x7b-DPO-inst & {69087} & {0.86$\pm$0.10} & {0.87$\pm$0.13}& {0.63$\pm$0.18}& {0.49$\pm$0.14} \\
        Mixtral-8x7b-DPO-inst & {72578} & {0.88$\pm$0.10} & {0.90$\pm$0.10}& {0.63$\pm$0.18}& {0.57$\pm$0.14} \\
        Mixtral-8x7b-DPO-inst & {72600} & {0.89$\pm$0.10} & {0.91$\pm$0.10}& {0.63$\pm$0.18}& {0.58$\pm$0.15} \\
        Mixtral-8x7b-DPO-inst & {72869} & {0.90$\pm$0.09} & {0.92$\pm$0.10}& {0.61$\pm$0.17}& {0.55$\pm$0.15} \\
        Mixtral-8x7b-DPO-inst & {72958} & {0.90$\pm$0.09} & {0.93$\pm$0.09}& {0.57$\pm$0.16}&{0.57$\pm$0.15} \\
        Mixtral-8x7b-DPO-inst & {72963} & {0.89$\pm$0.10} & {0.92$\pm$0.10}& {0.56$\pm$0.16}& {0.55$\pm$0.15} \\
        Mixtral-8x7b-DPO-inst & {72972} & {0.89$\pm$0.10} & {0.91$\pm$0.09}& {0.55$\pm$0.16}& {0.54$\pm$0.15} \\
        Mixtral-8x7b-DPO-inst & {73017} & {0.87$\pm$0.10} & {0.91$\pm$0.10}& {0.55$\pm$0.14}&{0.54$\pm$0.14} \\
        Mixtral-8x7b-DPO-inst & {73042} & {0.87$\pm$0.10} & {0.91$\pm$0.09}& {0.57$\pm$0.14}&{0.55$\pm$0.14} \\
        Mixtral-8x22b-inst & {69087} & {0.92$\pm$0.08} & {0.93$\pm$0.09}& {0.73$\pm$0.13}&{0.73$\pm$0.11} \\
        Mixtral-8x22b-inst & {72578} & {0.92$\pm$0.08} & {0.95$\pm$0.07}& {0.76$\pm$0.12}&{0.76$\pm$0.12} \\
        Mixtral-8x22b-inst & {72600} & {0.93$\pm$0.08} & {0.96$\pm$0.07}& {0.77$\pm$0.12}& {0.78$\pm$0.11} \\
        Mixtral-8x22b-inst & {72869} & {0.93$\pm$0.08} & {0.97$\pm$0.07}& {0.78$\pm$0.12}&{0.80$\pm$0.11} \\
        Mixtral-8x22b-inst & {72958} & {0.93$\pm$0.08} & {0.97$\pm$0.06}& {0.79$\pm$0.12}&{0.80$\pm$0.11} \\
        Mixtral-8x22b-inst & {72963} & {0.92$\pm$0.09} & {0.97$\pm$0.06}& {0.78$\pm$0.12}&{0.78$\pm$0.12} \\
        Mixtral-8x22b-inst & {72972} & {0.92$\pm$0.09} & {0.97$\pm$0.07}& {0.78$\pm$0.12}&{0.79$\pm$0.12} \\
        Mixtral-8x22b-inst & {73017} & {0.93$\pm$0.09} & {0.97$\pm$0.07}& {0.78$\pm$0.12}&{0.79$\pm$0.12} \\
        Mixtral-8x22b-inst & {73042} & {0.93$\pm$0.09} & {0.97$\pm$0.07}& {0.78$\pm$0.12}&{0.79$\pm$0.12} \\
        Mistral-v2-7b-inst & {69087} & {0.85$\pm$0.10} & {0.87$\pm$0.11}& {0.64$\pm$0.15}&{0.66$\pm$0.13} \\
        Mistral-v2-7b-inst & {72578} & {0.85$\pm$0.11} & {0.87$\pm$0.10}& {0.63$\pm$0.15}&{0.65$\pm$0.14} \\
        Mistral-v2-7b-inst & {72600} & {0.86$\pm$0.11} & {0.87$\pm$0.10}& {0.64$\pm$0.14}&{0.67$\pm$0.14} \\
        Mistral-v2-7b-inst & {72869} & {0.85$\pm$0.11} & {0.87$\pm$0.11}& {0.64$\pm$0.15}& {0.68$\pm$0.13} \\
        Mistral-v2-7b-inst & {72958} & {0.86$\pm$0.10} & {0.88$\pm$0.11}& {0.65$\pm$0.15}&{0.68$\pm$0.14} \\
        Mistral-v2-7b-inst & {72963} & {0.83$\pm$0.11} & {0.88$\pm$0.11}& {0.64$\pm$0.14}&{0.68$\pm$0.14} \\
        Mistral-v2-7b-inst & {72972} & {0.84$\pm$0.11} & {0.88$\pm$0.10}& {0.63$\pm$0.14}&{0.68$\pm$0.14} \\
        Mistral-v2-7b-inst & {73017} & {0.83$\pm$0.11} & {0.88$\pm$0.10}& {0.63$\pm$0.14}&{0.68$\pm$0.14} \\
        Mistral-v2-7b-inst & {73042} & {0.83$\pm$0.11} & {0.88$\pm$0.10}& {0.63$\pm$0.14}&{0.68$\pm$0.14}\\
        Mistral-v1-7b-inst & {69087} & {0.74$\pm$0.04}&{0.82$\pm$0.03}& /&/ \\
        Mistral-v1-7b-inst & {72578} & {0.75$\pm$0.04} & {0.81$\pm$0.03}& /&/ \\
        Mistral-v1-7b-inst & {72600} & {0.74$\pm$0.04} & {0.80$\pm$0.03}& /&/ \\
        Mistral-v1-7b-inst & {72869} & {0.74$\pm$0.04} & {0.81$\pm$0.03}& /&/ \\
        Mistral-v1-7b-inst & {72958} & {0.74$\pm$0.04} & {0.81$\pm$0.03}& /&/ \\
        Mistral-v1-7b-inst & {72963} & {0.74$\pm$0.04} & {0.80$\pm$0.03}& /&/ \\
        Mistral-v1-7b-inst & {72972} & {0.75$\pm$0.04} & {0.80$\pm$0.03}& /&/ \\
        Mistral-v1-7b-inst & {73017} & {0.74$\pm$0.04} & {0.80$\pm$0.03}& /&/ \\
        Mistral-v1-7b-inst & {73042} & {0.75$\pm$0.04} & {0.80$\pm$0.03}& /&/ \\
        Gemma-1.1-7b-inst & {69087} & {0.82$\pm$0.03} & {0.88$\pm$0.03}& /&/ \\
        Gemma-1.1-7b-inst & {72578} & {0.83$\pm$0.04} & {0.89$\pm$0.03}& /&/ \\
        Gemma-1.1-7b-inst & {72600} & {0.83$\pm$0.04} & {0.88$\pm$0.03}& /&/ \\
        Gemma-1.1-7b-inst & {72869} & {0.84$\pm$0.04} & {0.89$\pm$0.03}& /&/ \\
        Gemma-1.1-7b-inst & {72958} & {0.84$\pm$0.04} & {0.89$\pm$0.03}& /&/ \\
        Gemma-1.1-7b-inst & {72963} & {0.84$\pm$0.04} & {0.88$\pm$0.03}& /&/ \\
        Gemma-1.1-7b-inst & {72972} & {0.84$\pm$0.04} & {0.87$\pm$0.03}& /&/ \\
        Gemma-1.1-7b-inst & {73017} & {0.83$\pm$0.04} & {0.87$\pm$0.03}& /&/ \\
        Gemma-1.1-7b-inst & {73042} & {0.84$\pm$0.04} & {0.87$\pm$0.03}& /&/ \\
        GPT-3.5-turbo & {69087} & {0.86$\pm$0.03} & {0.88$\pm$0.03}& /&{0.69$\pm$0.04} \\
        GPT-3.5-turbo & {72578} & {0.87$\pm$0.03} & {0.89$\pm$0.03}& / &{0.69$\pm$0.04}\\
        GPT-3.5-turbo & {72600} & {0.87$\pm$0.03} & {0.89$\pm$0.03}& /&{0.67$\pm$0.04} \\
        GPT-3.5-turbo & {72869} & {0.87$\pm$0.03} & {0.90$\pm$0.03}& /&{0.67$\pm$0.04} \\
        GPT-3.5-turbo & {72958} & {0.87$\pm$0.03} & {0.90$\pm$0.03}& /&{0.67$\pm$0.04} \\
        GPT-3.5-turbo & {72963} & {0.86$\pm$0.03} & {0.89$\pm$0.03}& /&{0.67$\pm$0.04} \\
        GPT-3.5-turbo & {72972} & {0.86$\pm$0.03} & {0.88$\pm$0.03}& /&{0.67$\pm$0.04} \\
        GPT-3.5-turbo & {73017} & {0.85$\pm$0.03} & {0.88$\pm$0.03}& /&{0.67$\pm$0.04}\\
        GPT-3.5-turbo & {73042} & {0.85$\pm$0.03} & {0.88$\pm$0.03}& /&{0.67$\pm$0.04} \\
        \bottomrule
    \end{tabular}
    \label{tab:accuracy_diff_results_part2}
\end{table}

\newpage

\subsection{Results per book}
\begin{figure*}[htbp]
    \centering
    \begin{subfigure}[t]{0.5\textwidth}
        \centering
        \includegraphics[height=2in]{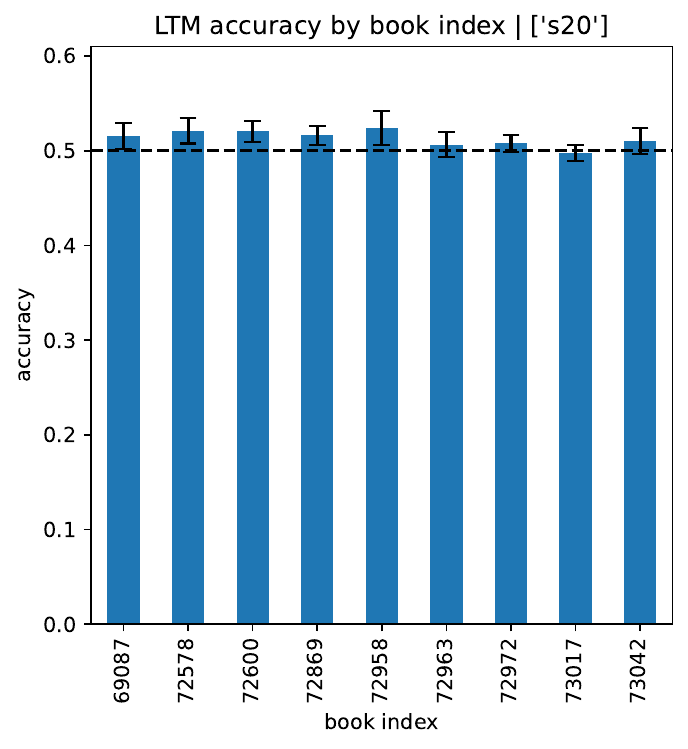}
        \caption{Segment length 20}
    \end{subfigure}%
    ~ 
    \begin{subfigure}[t]{0.5\textwidth}
        \centering
        \includegraphics[height=2in]{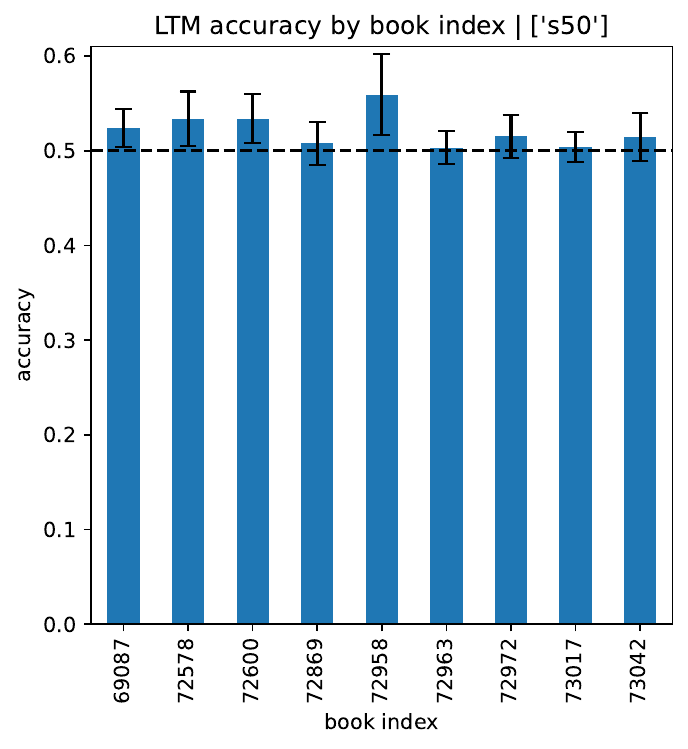}
        \caption{Segment length 50}
    \end{subfigure}
    \caption{Models' baseline performance by book (error bars indicate standard deviation)}
    \label{fig:LTM-by-book}
\end{figure*}

In Fig. \ref{fig:LTM-by-book}, we provide the baseline results without text-specific memory separately for each of the $9$ books in Book-SORT.

\begin{figure*}[htbp]
    \centering
    \begin{subfigure}[t]{0.5\textwidth}
        \centering
        \includegraphics[height=2in]{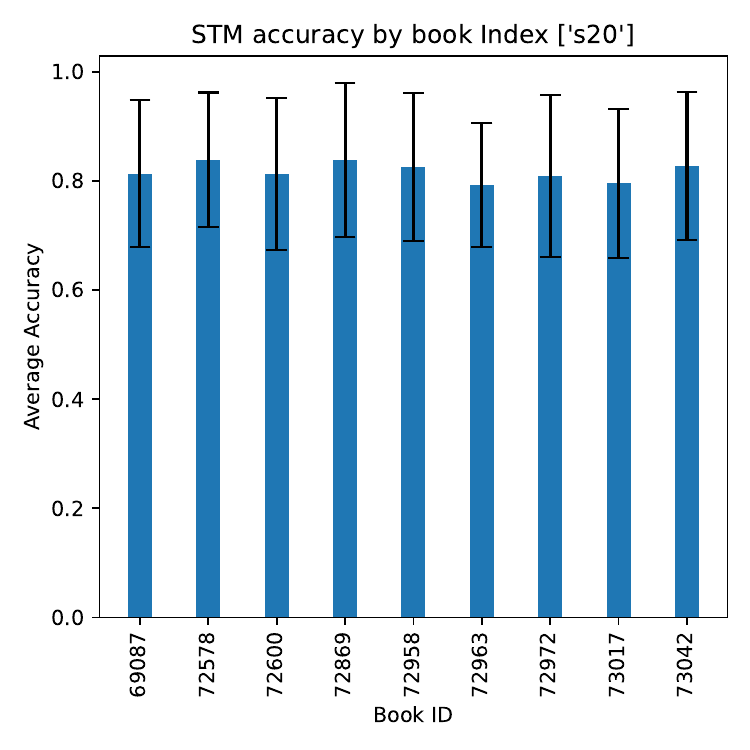}
        \caption{Segment length 20}
    \end{subfigure}%
    ~ 
    \begin{subfigure}[t]{0.5\textwidth}
        \centering
        \includegraphics[height=2in]{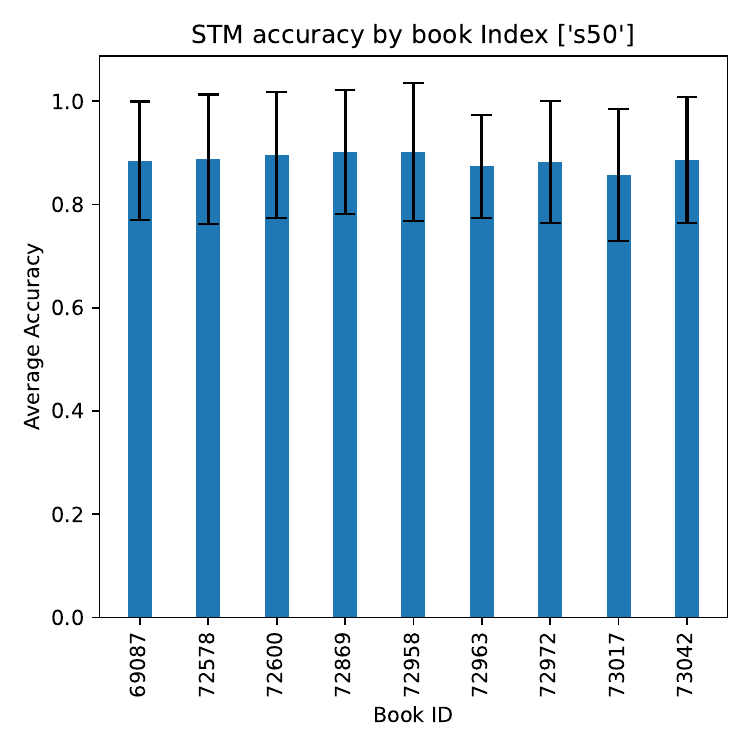}
        \caption{Segment length 50}
    \end{subfigure}
    \caption{Models' in-context memory performance by book (error bars indicate standard deviation)}
    \label{fig:STM-by-book}
\end{figure*}

In Fig. \ref{fig:STM-by-book}, we provide the in-context memory results separately for each of the $9$ books in Book-SORT.

\subsection{Relationship between in-context memory results and distance between segments across excerpt lengths}
\begin{figure*}[t]
    \begin{subfigure}[t]{1.0\textwidth}
        \centering
        \includegraphics[width=0.3\textwidth]{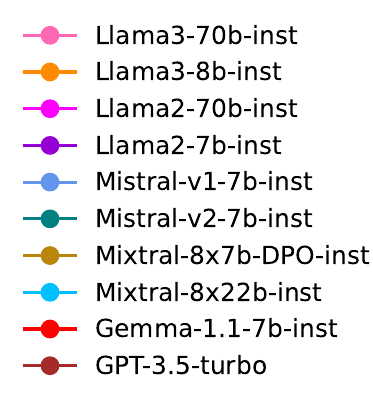}
    \end{subfigure}\\
    
    \begin{subfigure}[t]{0.5\textwidth}
    \includegraphics[height=2in]{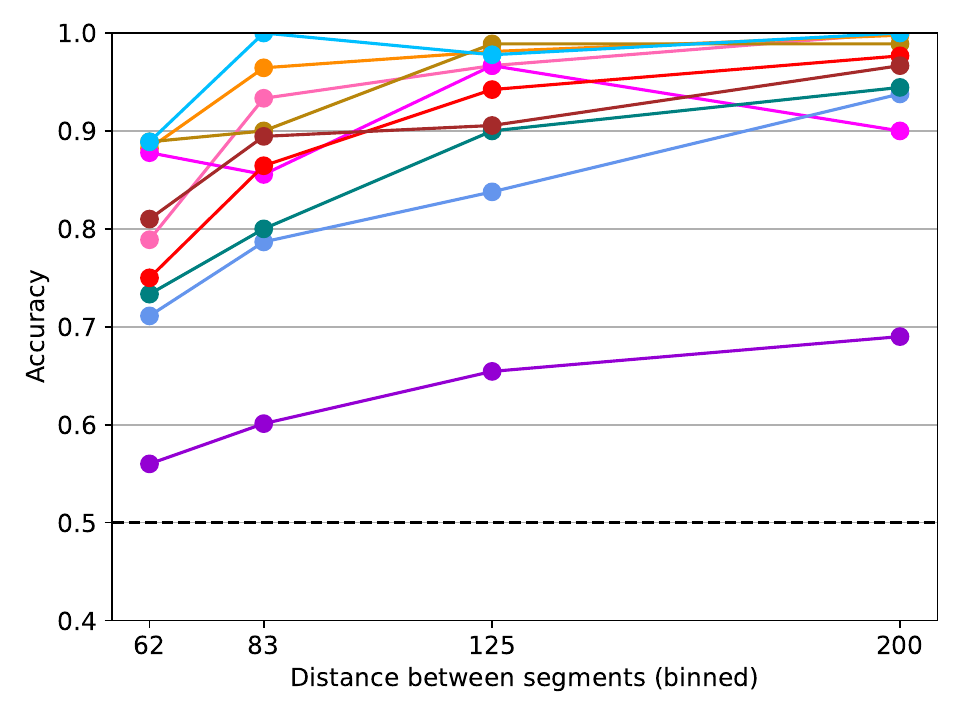}
        \caption{Excerpt length 250, segment length 20}
    \end{subfigure}%
    \begin{subfigure}[t]{0.5\textwidth}
    \includegraphics[height=2in]{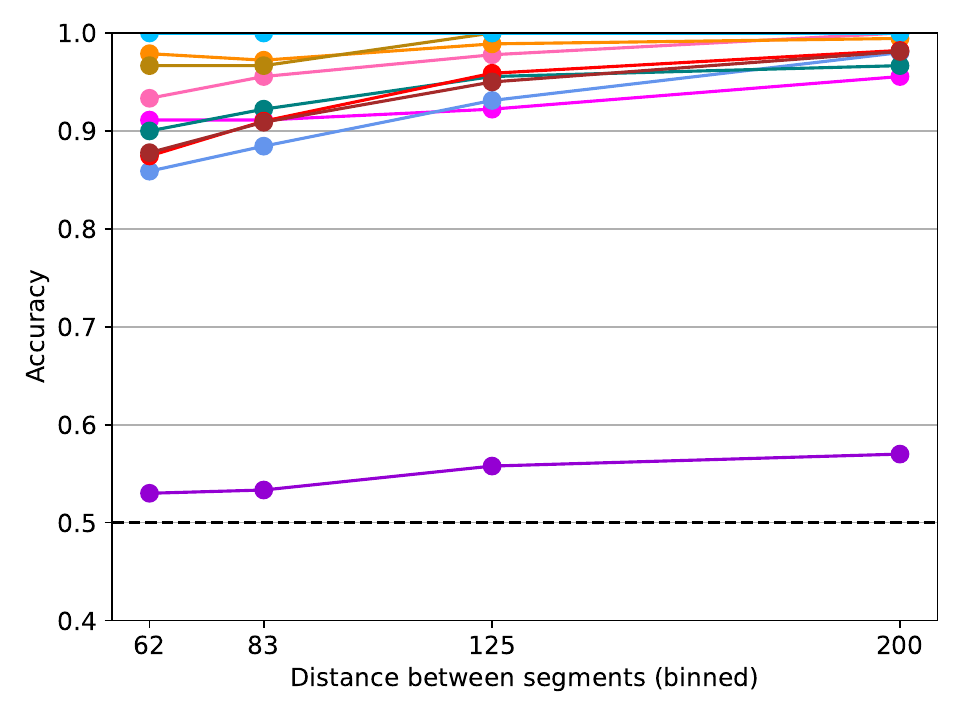}
        \caption{Excerpt length 250, segment length 50}
    \end{subfigure} \\

        \begin{subfigure}[t]{0.5\textwidth}
    \includegraphics[height=2in]{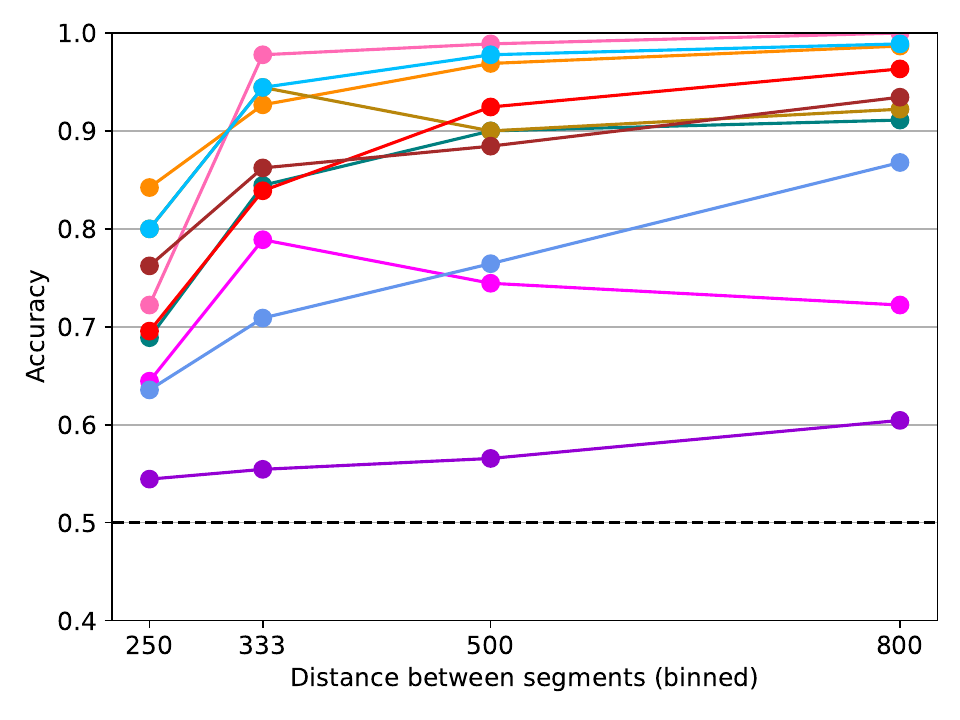}
        \caption{Excerpt length 1000, segment length 20}
    \end{subfigure}%
    \begin{subfigure}[t]{0.5\textwidth}
    \includegraphics[height=2in]{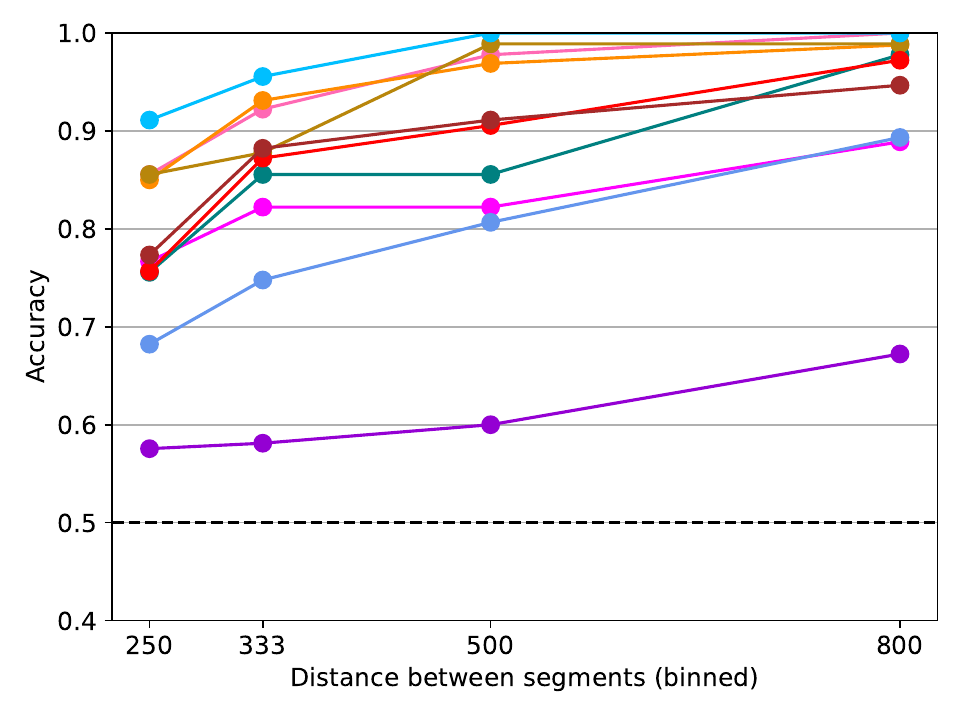}
        \caption{Excerpt length 1000, segment length 50}
    \end{subfigure} \\

    \begin{subfigure}[t]{0.5\textwidth}
    \includegraphics[height=2in]{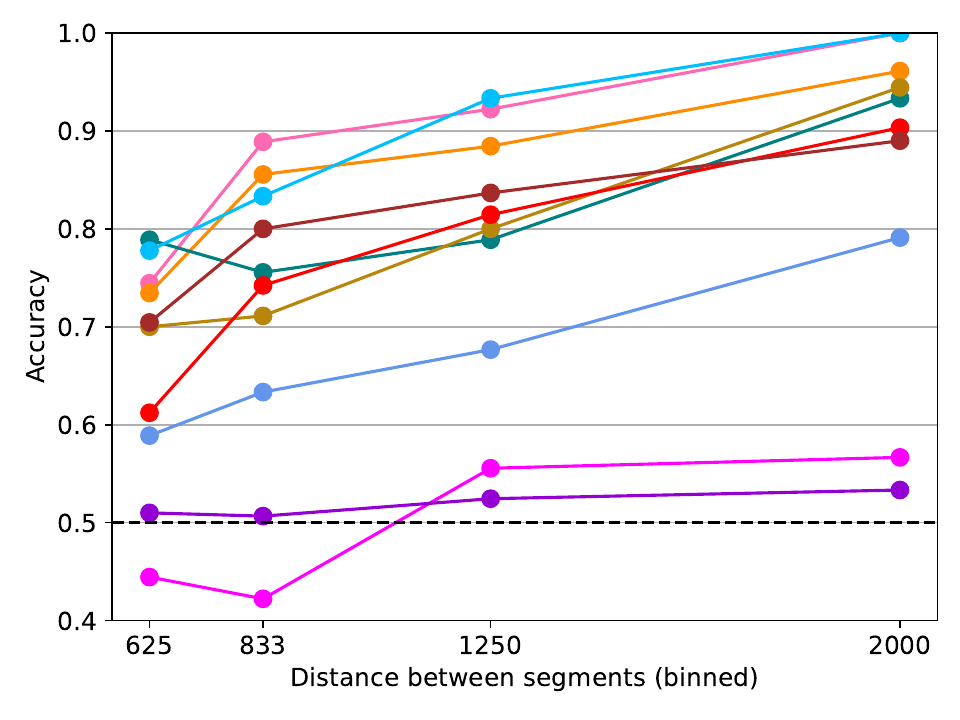}
        \caption{Excerpt length 2500, segment length 20}
    \end{subfigure}%
    \begin{subfigure}[t]{0.5\textwidth}
    \includegraphics[height=2in]{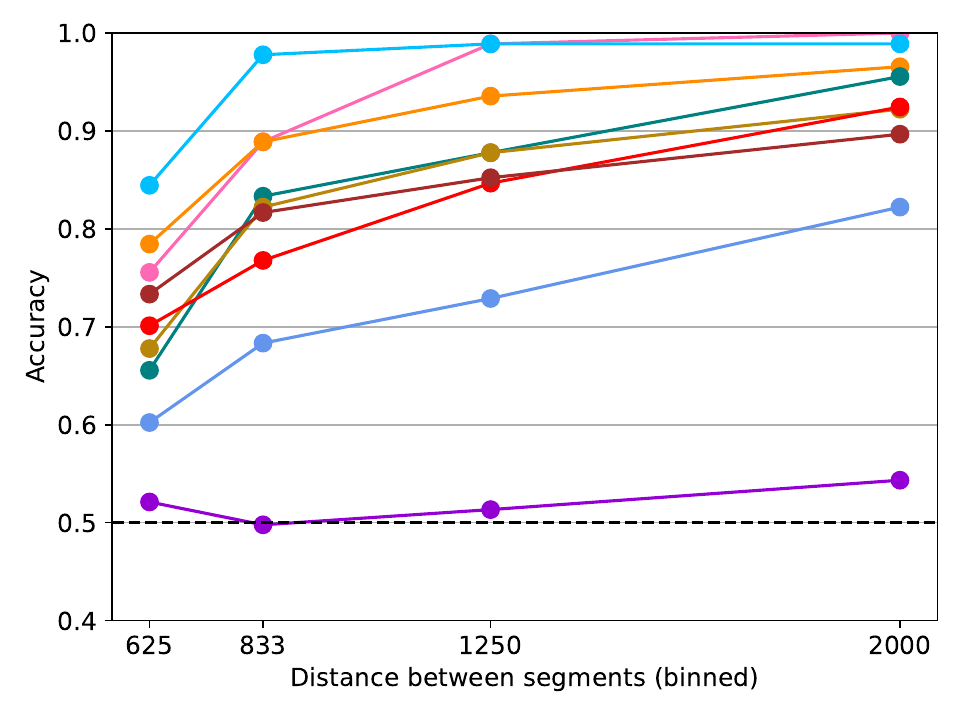}
        \caption{Excerpt length 2500, segment length 50}
    \end{subfigure} \\
    
    \caption{Average accuracy by distance between segments (All excerpt length), part A.}
    \label{fig:stm_bins_1}
\end{figure*}

\begin{figure*}[t]
    \begin{subfigure}[t]{1.0\textwidth}
        \centering
        \includegraphics[width=0.3\textwidth]{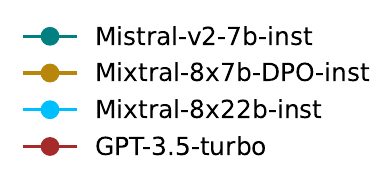}
    \end{subfigure}\\
    \begin{subfigure}[t]{0.5\textwidth}
    \includegraphics[height=2in]{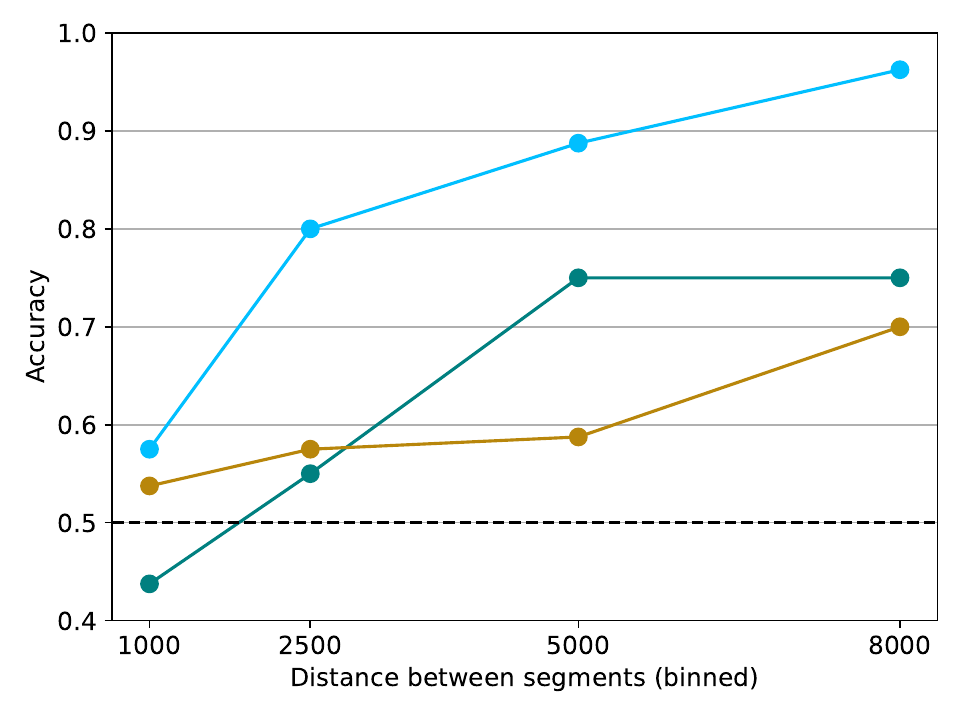}
        \caption{Excerpt length 10000, segment length 20}
    \end{subfigure}%
    \begin{subfigure}[t]{0.5\textwidth}
    \includegraphics[height=2in]{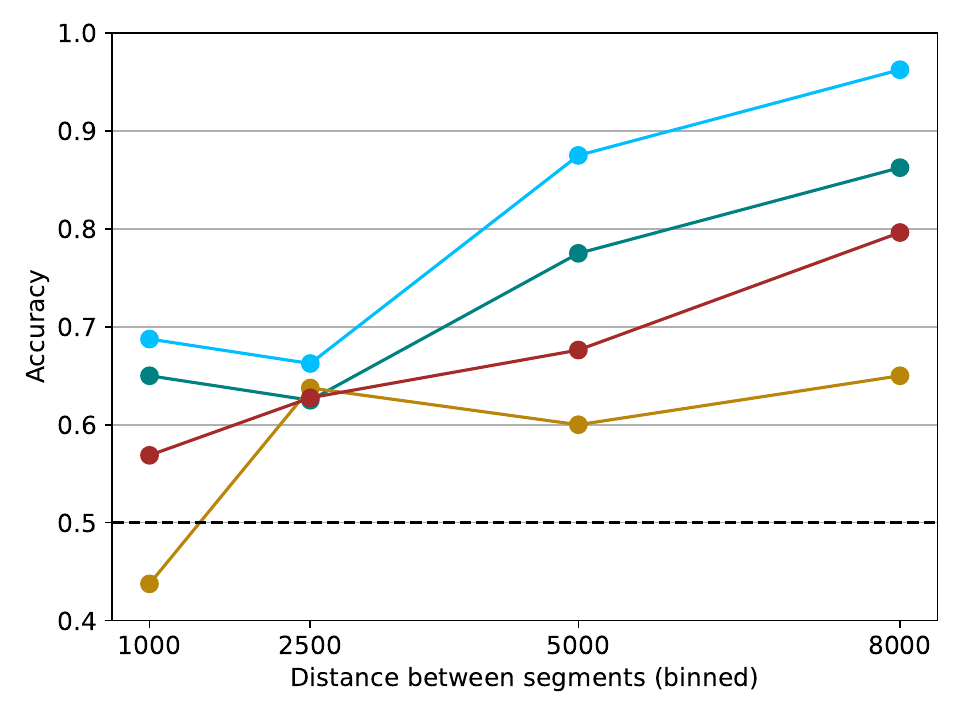}
        \caption{Excerpt length 10000, segment length 50}
    \end{subfigure} \\

        \begin{subfigure}[t]{0.5\textwidth}
    \includegraphics[height=2in]{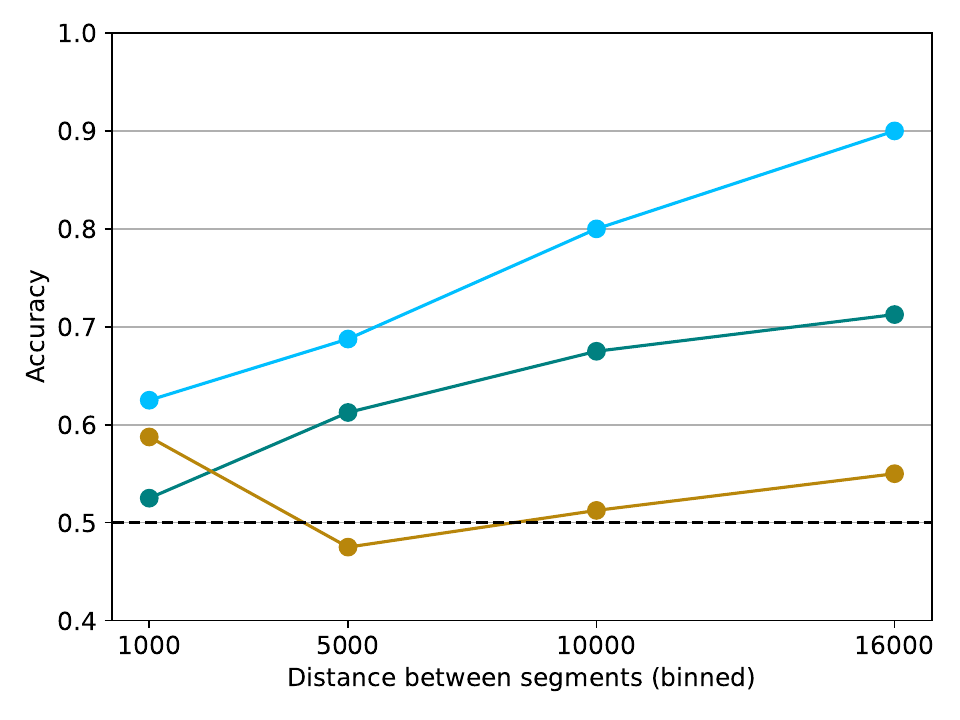}
        \caption{Excerpt length 20000, segment length 20}
    \end{subfigure}%
    \begin{subfigure}[t]{0.5\textwidth}
    \includegraphics[height=2in]{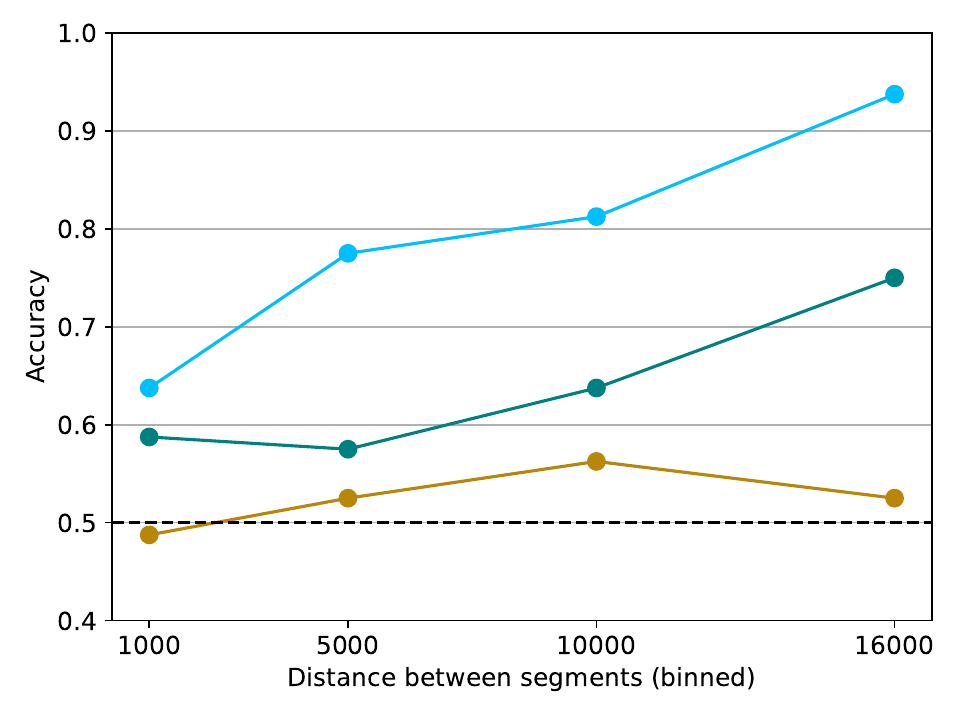}
        \caption{Excerpt length 20000, segment length 50}
    \end{subfigure} \\
    
    \caption{Average accuracy by distance between segments (All excerpt length), part B.}
    \label{fig:stm_bins_2}
\end{figure*}

In Fig. \ref{fig:stm_bins_1} and Fig \ref{fig:stm_bins_2}, we show the average accuracy by the distance between segments for all the excerpt lengths and segment lengths.

\subsection{Baseline performance}
\begin{figure*}[t]
    \begin{subfigure}[t]{1.0\textwidth}
        \centering
        \includegraphics[width=1\textwidth]{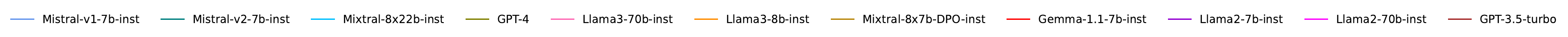}
    \end{subfigure}\\
    \begin{subfigure}[t]{0.5\textwidth}
        \centering
        \includegraphics[height=2in]{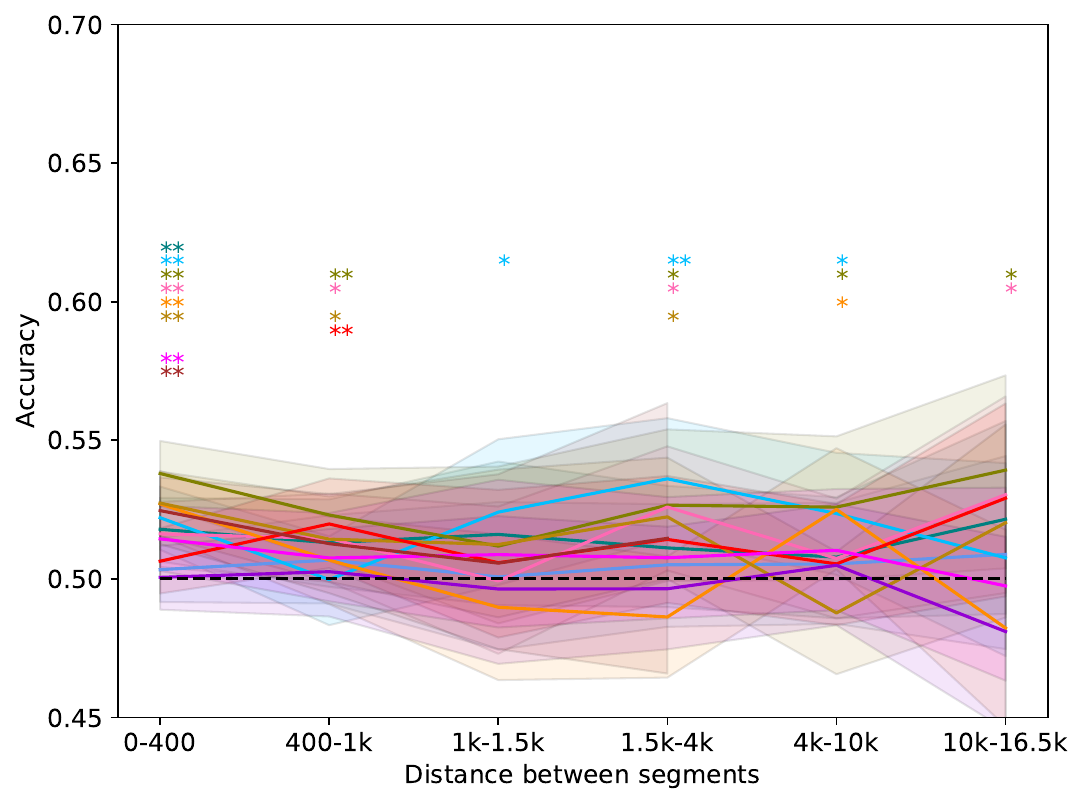}
        \caption{Segment length 20}
    \end{subfigure}%
    \begin{subfigure}[t]{0.5\textwidth}
        \centering
        \includegraphics[height=2in]{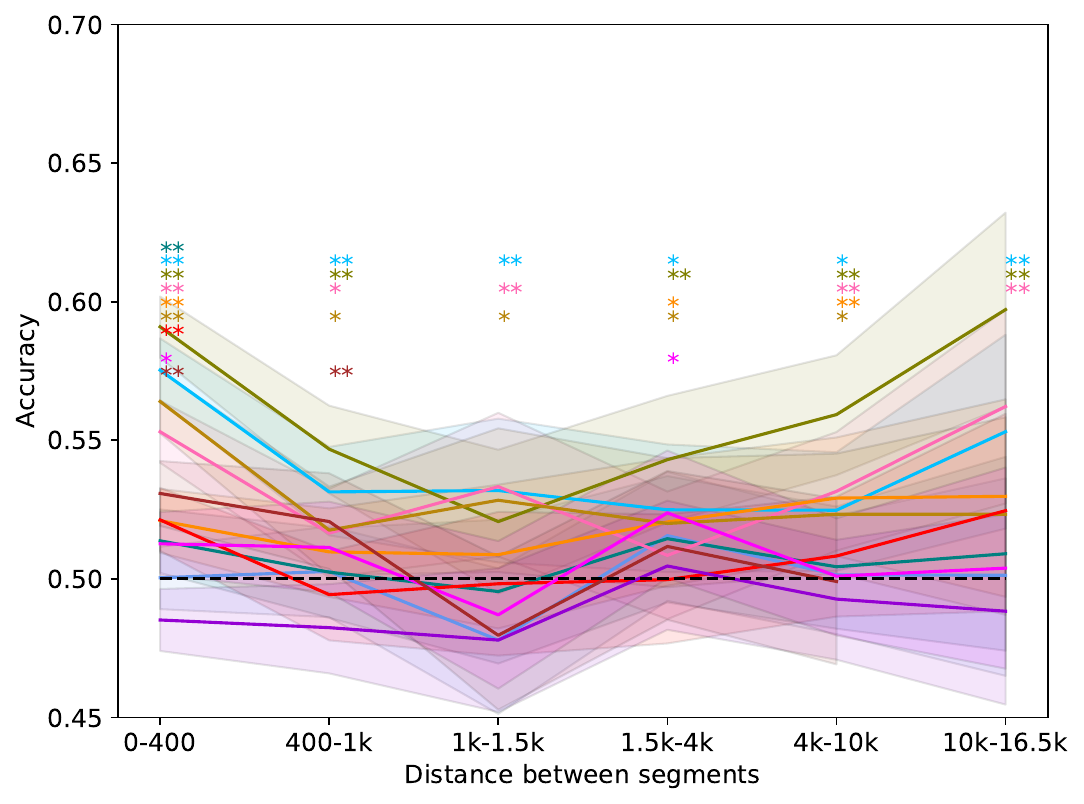}
        \caption{Segment length 50}
    \end{subfigure}
    \caption{Baseline model performance on SORT without text-specific memory by segment distance (95\% bootstrapped confidence interval). Significant difference from chance is marked with asterisks ($^*$p-value$<$0.05,$^{**}$p-value$<$0.01).}
    \label{appendix:ltm-allmodels}
\end{figure*}

In Fig. \ref{appendix:ltm-allmodels}, we provide the SORT results based on parametric memory for all models across various segment distances. Due to the recent addition of the texts in Book-SORT to the public domain, we expect that models were not trained on these texts, i.e. they should not have text-specific memory. Performance is higher for segment pairs that have a short distance and a high distance in the books, indicating that these are more likely to be sort-able without episodic memory, based on temporal order reasoning.

\section{\datasetname$ $ results from additional models}
\label{appendix:more-models}
\subsection{Base models}
\begin{figure*}[htbp]
    \centering
    \begin{subfigure}[t]{0.5\textwidth}
        \centering
        \includegraphics[height=2in]{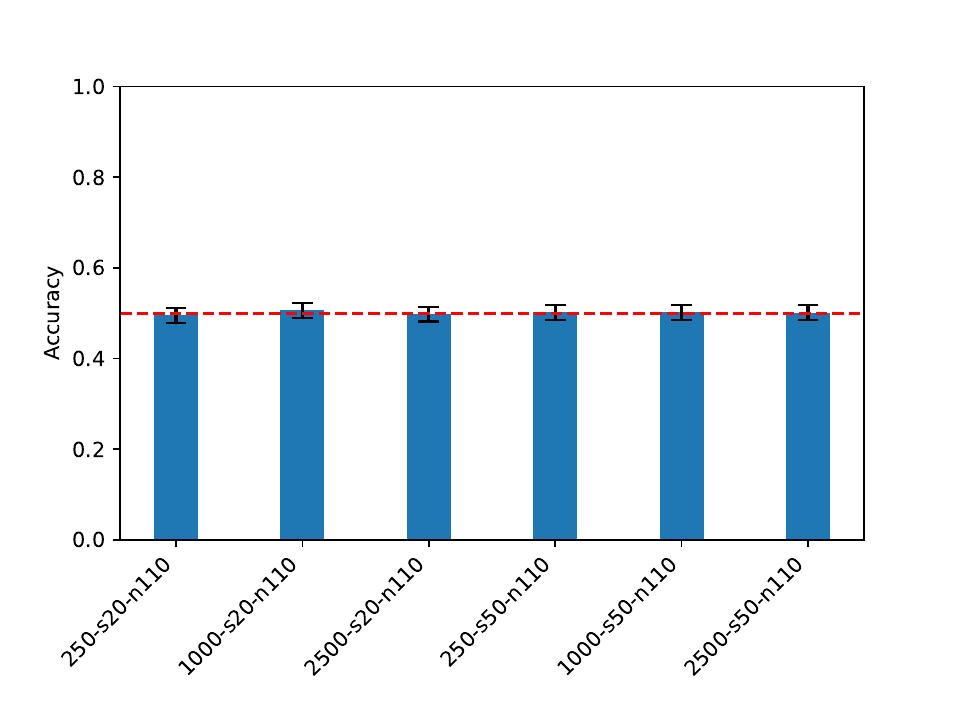}
        \caption{Llama3-8b}
    \end{subfigure}%
    ~ 
    \begin{subfigure}[t]{0.5\textwidth}
        \centering
        \includegraphics[height=2in]{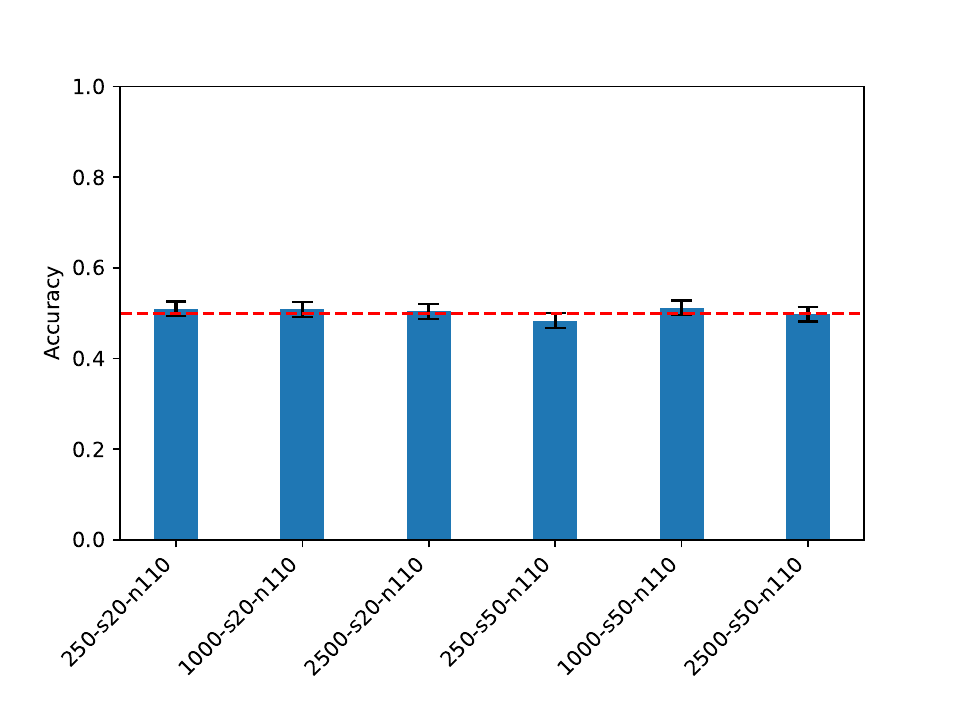}
        \caption{Mistral-7b}
    \end{subfigure}
    \caption{Base model performance for SORT (in-context memory).}
    \label{fig:base-stm}
\end{figure*}

We chose 2 base models to evaluate, Llama3-8b and Mistral-7b, whose fine-tuned versions (Llama3-8b-inst and Mistral-v2-7b-inst) performed well on SORT based on in-context memory. Figure \ref{fig:base-stm} shows that both the base models got around chance performance across all the excerpt lengths and segment lengths.

\subsection{State-space models}
We tested an instruction-tuned version of the state space model RWKV~\citep{peng-etal-2023-rwkv}, available in Huggingface as RWKV/rwkv-raven-7b. The results of the prompt sweep on SORT with in-context memory yielded a performance of 51\% -- very close to chance levels. A possibility for this is a larger sensitivity to prompting, e.g. this model might require instructions to be given in a different order. We assume that this is due to insufficient instruction tuning. While it could be interesting to see the performance of a state-space model with memory other than in-context, we leave this question to future work. 

\section{Finetuning of Llama3-8b-Instruct}
\label{appendix:finetuning}
\paragraph{Fine-tuning details.}

We fine-tuned Llama3-8b-Instruct and Mistral-7b-v0.2-Instruct on a single node with 8 A100 GPUs. The books (without pre-processing beyond removing Project Gutenberg related text, i.e. including chapter signifiers) are split into chunks of 5000 words and contextualized in the same way in which excerpts are presented in-context in our experiments, i.e. together with the book-title in a user prompt along with a preceding system prompt. For the instruction data, we exclude the following task types: "experience", "stylized\_response", "joke", "trivia", "roleplay", "riddle" and "greeting". Samples containing both book-chunks and instruction-following examples are padded to the maximum length in a batch. The effective batch size in our experiments is 192. We choose a moderately low initial learning rate of 5e-6 with cosine decay and a small amount of weight decay set to 1e-4. The chunks of books comprise a total of 116 independent samples. Together with 3\,500 instruction samples from the OpenHermes dataset \citep{OpenHermes25}, this means 19 steps of gradient descent are taken in one epoch. We fine-tuned both models for a total of 5 epochs.

\paragraph{Inclusion of instruction data to avoid catastrophic forgetting.} Fine-tuning an instruction-tuned model on specific data can lead to catastrophic forgetting \citep{luo2024empiricalstudycatastrophicforgetting}, such that only a few steps of gradient descent can be enough to undo previous behavioral alignment \citep{qi2023finetuning,zhan2024removing}. To retain the general ability to follow instructions, and to allow for control condition fine-tuned models in which the book text is not part of the training data, we include $3,500$ instruction samples from the OpenHermes2.5 dataset on Huggingface \citep{OpenHermes25} (see Appendix \ref{appendix:finetuning} for details). Therefore the baseline without text-specific memory to compare with is not only the respective initial model before fine-tuning, but the same model fine-tuned on the same $3,500$ instruction samples but excluding the 116 samples of book chunks.

\subsection{Perplexity analysis of fine-tuned models}
\label{appendix:perplexity_analysis}
\begin{figure}[h]
    \centering
    \includegraphics[width=0.75\linewidth]{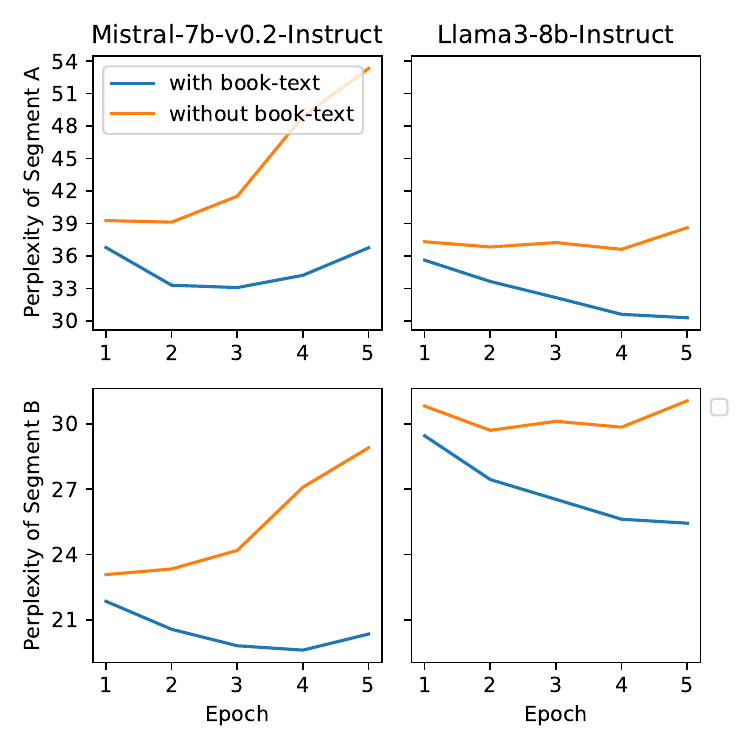}
    \caption{Perplexity of the two segments after fine-tuning of Mistral-7b-v0.2-Instruct and Llama3-8b-Instruct, when presented in the absence of in-context access to source excerpts.}
    \label{fig:finetuned-segment-perplexity}
\end{figure}

To confirm that fine-tuning on the books makes a model learn about the segments, we compare the perplexities of the two segments shown in SORT without source text presented in-context. We find that when the models are finetuned on data that includes the chunks of the books, they have a substantially lower perplexity for both segments, compared with the models fine-tuned only on the instruction data (see figure \ref{fig:finetuned-segment-perplexity}). Note that the scale of these perplexity values highlights that our task is likely out of distribution, presumably with little to no similar instruction data seen during pre-training and fine-tuning.

\subsection{Comparison of SORT performance after fine-tuning using McNemar's Test}
\label{appendix:mc_nemar}
\begin{figure}[h]
    \centering
    \includegraphics[width=0.75\linewidth]{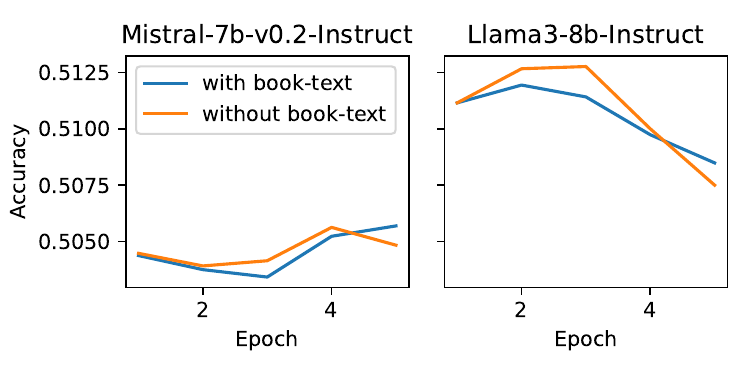}
    \caption{Accuracy of Llama3-8b-Instruct and Mistral-7b-v0.2-Instruct across epochs of finetuning on data including and excluding relevant book-text. Figure \ref{fig:mcnemar-matrix} shows that differences between accuracies shown here are not statistically significant (p>0.05).}
    \label{fig:accuracy_finetuning}
\end{figure}

\begin{figure}[t]
    \begin{subfigure}[t]{0.5\textwidth}
    \includegraphics[width=\textwidth]{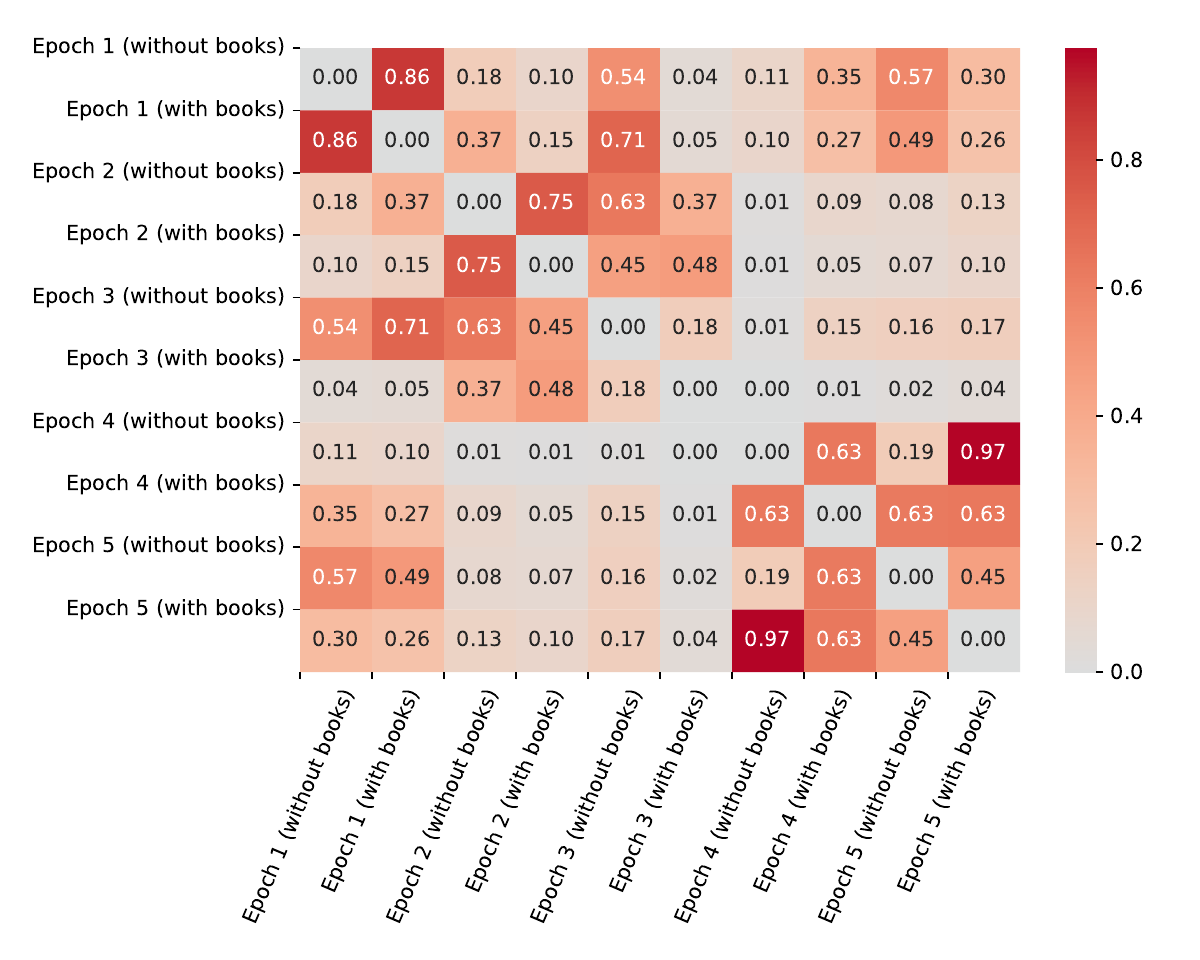}
        \caption{Mistral-7b-v0.2-Instruct}
        \label{fig:mcnemar-mistral}
    \end{subfigure}
    \begin{subfigure}[t]{0.5\textwidth}
    \includegraphics[width=\textwidth]{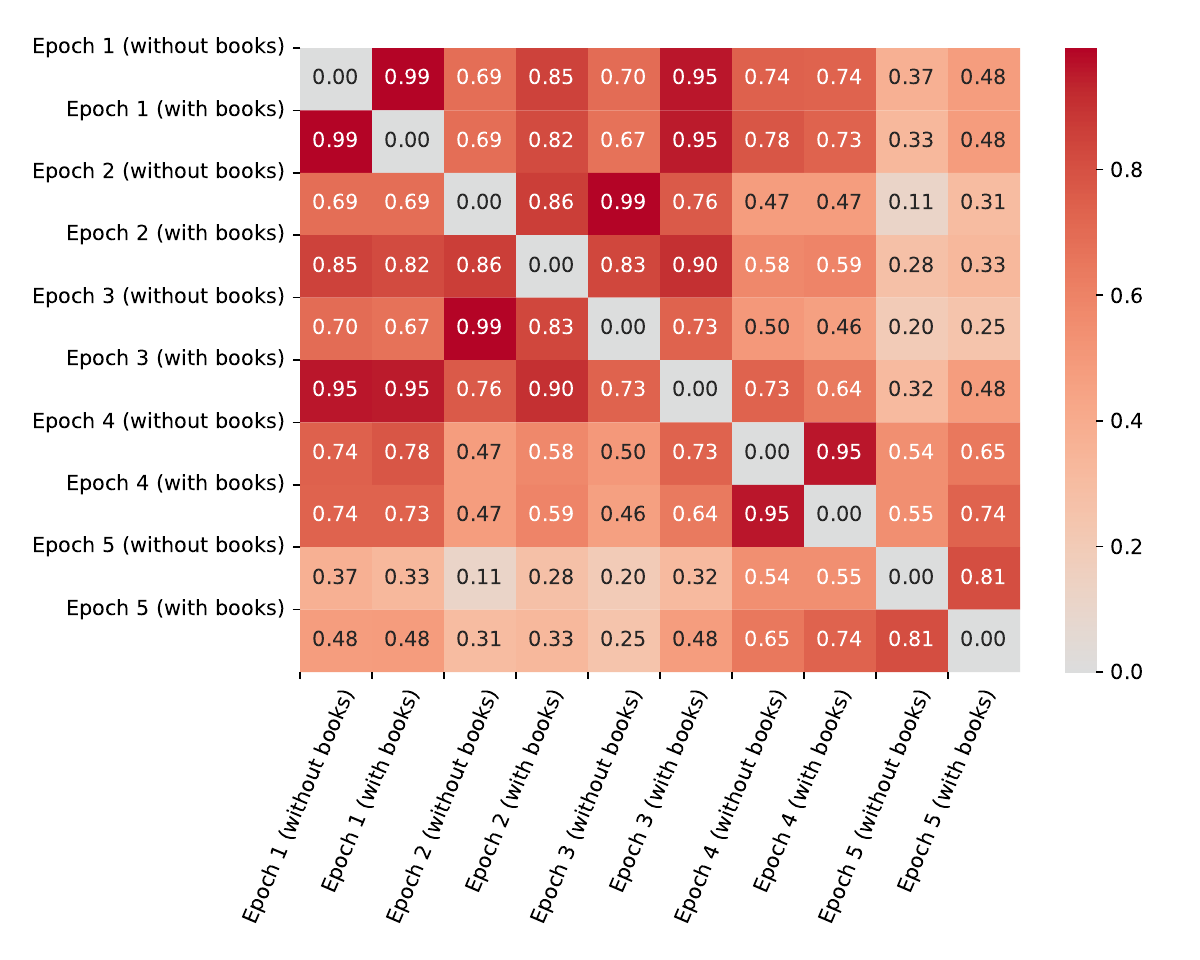}
        \caption{Llama3-8b-Instruct}
        \label{fig:mcnemar-llama}
    \end{subfigure}
    \caption{McNemar's Test matrix of fine-tuned models performance. Shown are p-values indicating whether a model checkpoint (row) is different in its accuracy compared to another checkpoint (columns) with statistical significance. We fine-tuned with and without the books used in Book-SORT. There is no statistically significant difference between the models finetuned without and with book text. The effect of fine-tuning seems insignificant even without correcting these p-values for multiple comparisons.}
    \label{fig:mcnemar-matrix}
\end{figure}

 We find that even though the book-text finetuned Llama3-8b model has a form of memory of the books' texts, the epoch-matched performance between the models fine-tuned without the book-chunks does not differ statistically for any epoch (Figure \ref{fig:accuracy_finetuning}). For this analysis we use McNemar's test since we have an exact match of presented samples for both the memory-finetuned model and the baseline that does not form any memory of the text (Figure \ref{fig:finetuned-segment-perplexity}). We find high p-values, indicating no difference in performance between models fine-tuned with and without the book text (Figure \ref{fig:mcnemar-matrix}), neither for Llama3-8b-Instruct, nor for Mistral-7b-v0.2-Instruct.
 
\subsection{Comparison of SORT performance after fine-tuning using a pairwise t-test}
\label{appendix:t-test}

Testing the binary correctness evaluated based on a greedily sampled token does not allow us to draw conclusions about sub-threshold effects of fine-tuning on task performance.
To test whether the models fine-tuned on the books is better than the models that are fine-tuned without chunks from the books, we performed a pairwise t-test on a continuous measure of accuracy based on the token log-probabilities. We compute the likelihood of the correct answer by taking the log ratio of the correct answer among all answers that can be mapped to either A or B, i.e.  we are interested in $\text{log}\left(\frac{p(a=y)}{(p(a=A)+p(a=B)}\right)$, where $y$ is the correct answer.

\begin{figure}[t]
    \begin{subfigure}[t]{0.5\textwidth}
    \includegraphics[width=\textwidth]{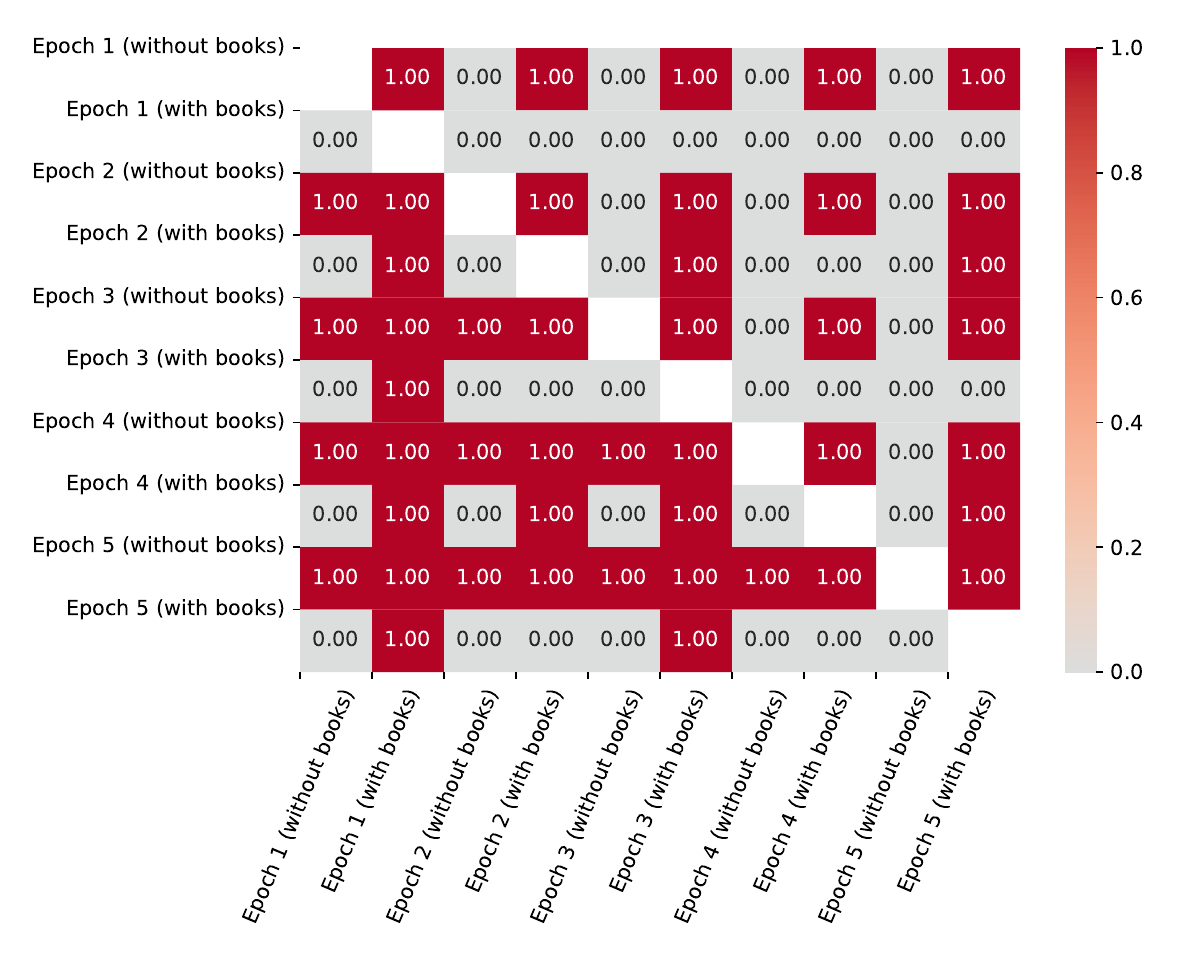}
        \caption{Mistral-7b-v0.2-Instruct}
        \label{fig:ttest-mistral}
    \end{subfigure}
    \begin{subfigure}[t]{0.5\textwidth}
    \includegraphics[width=\textwidth]{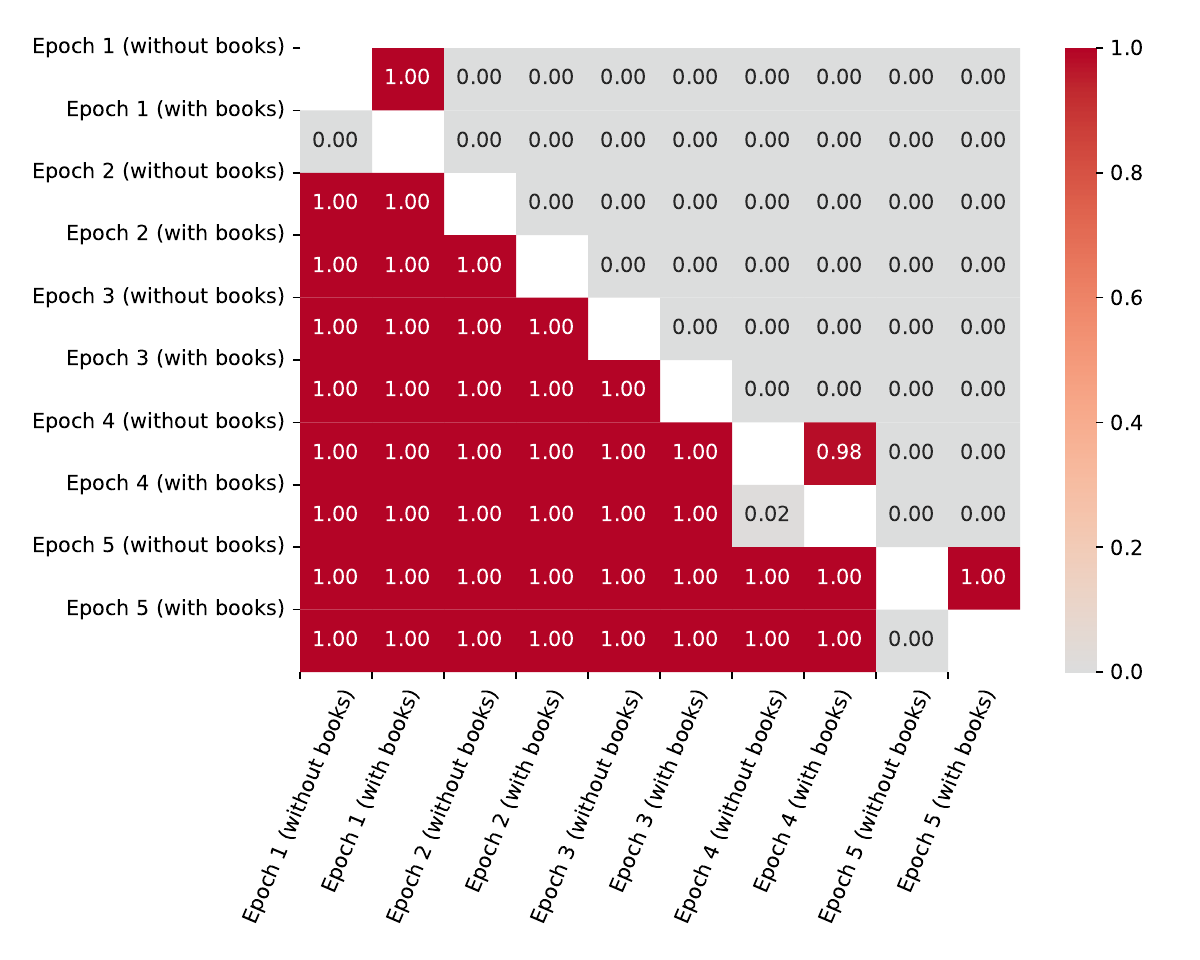}
        \caption{Llama3-8b-Instruct}
        \label{fig:ttest-llama}
    \end{subfigure}
    \caption{Pairwise t-test matrix of fine-tuned models. Shown are p-values indicating whether a model (row) has higher log probabilities of the correct answer compared to another model (columns) with statistical significance.}
    \label{fig:t-test-matrix}
\end{figure}

\begin{figure}[t]
    \centering
    \includegraphics[width=1\linewidth]{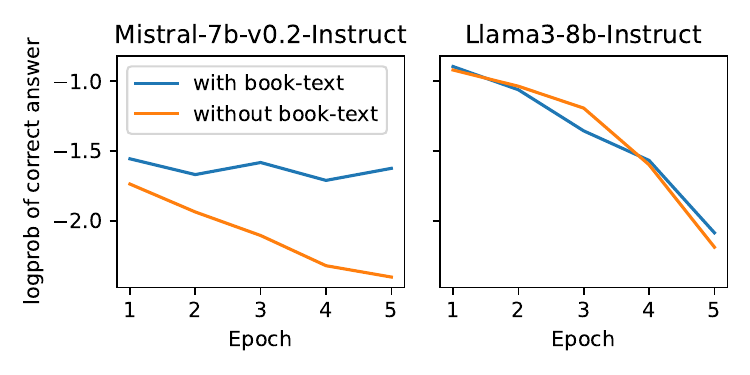}
    \caption{Log-probability of the correct answer for fine-tuned models across epochs. Figure \ref{appendix:t-test} shows statistical significance between conditions and epochs for this data.}
    \label{fig:logprob_change_finetuning}
\end{figure}

The results shown in figure \ref{fig:t-test-matrix} suggest that fine-tuned models do improve over the base model, with the book text condition performing better than the others after one epoch of training with statistical significance ($p<0.01$). Even though there is an effect, the magnitude is very small, as can be seen in Figure \ref{fig:logprob_change_finetuning}, and this positive effect could also be attributed to interleaving the instruction data with samples including longer texts ($5,000$ words) compared to just the instruction samples.

\subsection{In-context memory performance of fine-tuned models}

Despite the inclusion of instruction data in fine-tuning, the accuracy with source excerpts presented in-context of SORT decreased from 0.93 to $0.90$ after a single epoch and to $0.88$ after three epochs of fine-tuning for Llama3-8b-Instruct. For the instruction-data only baseline of Llama3-8b-Instruct, the performance degraded slightly less with an accuracy of $0.91$ after the first epoch of fine-tuning.

\section{Code and Data}

We provide the code to create SORT datasets and evaluate models on SORT in a \href{https://github.com/bridge-ai-neuro/SORT}{public GitHub repository}. 
Our evaluation code currently supports the OpenAI API, Huggingface Transformers \citep{wolf2020huggingfaces} and vLLM \citep{kwon2023efficient} for distributed inference. Our initial Book-SORT dataset can be accessed via \href{https://huggingface.co/datasets/memari/booksort}{Huggingface Datasets}.

\paragraph{License.} We make our code and data openly available under a permissive BSD-3 license for code. Data including Book-SORT is available under a CC0 license.

\end{document}